\documentclass[times, review, 10pt]{elsarticle}

\usepackage{amssymb}
\usepackage{amsmath}
\usepackage{booktabs}
\usepackage{multirow}
\usepackage{graphicx}
\usepackage{bbm}
\usepackage{xcolor}
\usepackage{enumitem}
\usepackage{hyperref}

\newcommand{\revision}[1]{#1}

\journal{Pattern Recognition}

\begin{document}

\begin{frontmatter}



\title{Uncertainty Estimation in Instance Segmentation of
Affordances via Bayesian Visual Transformers} 


\author[label1]{Lorenzo Mur-Labadia}
\author[label1]{Ruben Martinez-Cantin}
\author[label1]{Jose J.Guerrero} 

\affiliation[label1]{organization={I3A, Universidad de Zaragoza},
            city={Zaragoza},
            postcode={50018}, 
            country={Spain}}

\begin{abstract}
Visual affordances identify regions in an image with potential interactions, offering a novel paradigm for scene understanding. 
Recognizing affordances allows autonomous robots to act more naturally, could enhance human-robot interactions, enrich augmented reality systems, and benefit prosthetic vision devices. 
Accurate and localized prediction of affordance regions, rather than general saliency maps is crucial for these applications. 
We present a model for instance segmentation of affordances  by adopting sample-based and ensembles approaches for uncertainty estimation. We extend an attention-based architecture for our novel task, showing with detailed ablation experiments the effects of each component. By comparing the distribution of these different detections, we extract pixel-wise epistemic and aleatoric variances at both the semantic and spatial levels. 
In addition, we propose a novel measure called Probability-based Mask Quality, which enables a comprehensive analysis of semantic and spatial variations in a probabilistic instance segmentation model. 
Our results show that the global consensus of multiple sub-networks of Bayesian models improve deterministic networks due to a better mask refinement and generalization. 
This fact, joined with the more powerful features extracted by attention-based mechanisms, represent an improvement of +7.4 p.p on the $F_{\beta}^w$ score in the challenging IIT-Aff dataset. 
Bayesian models are also better calibrated, producing less overconfident probabilities and with a better uncertainty estimation. Qualitative results show that aleatoric variance appears in the contour of the objects, while the epistemic variance is observed in visual challenging pixels, adding interpretability to the neural network.
\end{abstract}





\end{frontmatter}











\section{Introduction}

The affordances visual perception identifies the regions on the image with potential interaction between the agent and the object, enhancing the scene understanding. Rooted in Gibson's foundational theory of perception, affordances are the potential actions that the environment offers to the agent based on its motor capabilities \cite{gibson1977theory}. 
When transferred to autonomous robots \cite{murphy1999case}, this knowledge would make them interact more naturally and obtain better grasping capabilities. 
In Augmented Reality (AR) systems, leveraging actions learned from expert demonstrations enhances user experience by providing context-specific guidance \cite{grauman2023ego}. For instance, an AR assistive device could illuminate parts of a new engine model for a mechanic, indicating the appropriate procedural actions, thereby facilitating a more intuitive repair process. 
Similarly, the constrained resolution of visual assistive devices for the visually impaired \cite{sanchez2020semantic} necessitates the isolation of specific regions indicative of potential user actions. Such applications require a detailed scene understanding that prioritizes localized affordance regions compared with diffuse saliency maps \cite{nagarajan2019grounded}. 
In this work, we segment the object boundaries and classify the affordable action associated with each object part, obtaining pixel-wise precision with a detailed estimation of the spatial and semantic uncertainty.



Recent advances in deep learning have enhanced visual perception: object detection \cite{he2017mask, doherty2025bifpn},  action anticipation \cite{furnari2019would} or affordances detection \cite{nagarajan2020ego} are some examples of the improvements in this area. However, despite the significant advances in performance, deep neural networks are overconfident sensors with low interpretability, producing misclassified predictions with high confidence \cite{guo2017calibration}. In this work, we aim to overcome these deficiencies with Bayesian deep learning alternatives, which produce better-calibrated probabilities with uncertainty estimation, obtaining a more interpretable and robust perception \cite{morilla2023robust}.

Uncertainty estimation in the problem of \emph{instance segmentation of affordances} aims to accurately segment object parts, classify them based on potential interactions, and quantify uncertainty for a more robust perception. This process is illustrated in Figure 1.
Our Bayesian detection model, inspired in Mask-RCNN \cite{he2017mask} is extended with intermediate sampling layers for enhanced performance and uncertainty quantification. During inference, our sampling-based approach executes $N$ forward passes, generating multiple predictions—termed detections—from the same input, thereby enhancing the model's reliability. The detections are then clustered based on their spatial and semantic affinity in observations. Comparing the distribution of the grouped observations, we extract the epistemic and aleatoric variance both at spatial and semantic levels. The spatial uncertainty is associated with the binary probability of having a mask around an object, thus revealing pixel-level spatial uncertainty. Semantic uncertainty is derived from the variability within the class probability vectors, highlighting discrepancies in assigning an afforded action to the object part.

This paper is an extension of our previous work \cite{aff_loren} with the following contributions. First, we substitute the previous convolutional-based architecture with a attention-based backbone \cite{liu2021swin}. Second, we compare different techniques for uncertainty estimation: Monte-Carlo Dropout (MC-Dropout)\cite{gal2016dropout}, Mask Ensembles (Mask-Ens) \cite{durasov2021masksembles}, Deep Ensembles (Deep-Ens) \cite{lakshminarayanan2017simple}, and Snapshot Ensembles (Snap-Shot) \cite{huang2017snapshot}. Third, we design the novel Probability-based Mask Quality (PMQ) metric that evaluates the uncertainty estimation of the predictions. We achieve a new state-of-the-art at 90.6 $\%$ on the $F_{\beta}^w$ score, which represents +7.4 percentage points of improvements compared with previous works in the IIT-Aff dataset. We report detailed ablation studies on the Bayesian techniques and the disposition of the sampling layers within the architecture, showing the benefits of Bayesian models over their respective deterministic versions. The reported semantic and spatial ECE and AUSE metrics show that the calibration of our attention-based backbone is better than their homologous convolutional versions. Our qualitative results details the uncertainty estimation reflected on the epistemic and aleatoric contributions of the spatial and semantic variance, showing a more interpretable perception.

\section{Related works}


\subsection{Visual perception of affordances}

Gibson's psychological theory of affordances \cite{gibson1977theory} has inspired multiple computer vision works due to its applications in robotics \cite{koppula2015anticipating} or scene understanding \cite{mur2023multi}. 
Koppula et al. \cite{koppula2015anticipating} anticipate future human activities through object affordances. The work by Montesano et al. \cite{affmontesano} addresses the learning of affordances through robot-environment interactions using a Bayesian network model. Detailed affordance masks are exploited by Yang et al. \cite{yang2023watch} for learning a better robot manipulation by selecting accurately between \textit{pickable} and \textit{placeable} objects. Ego-Topo \cite{nagarajan2019grounded} constructed an affordance topological obtaining activity-centric zones on the nodes and used this representation for long-term activity recognition, showing the potential of affordances in video understanding. Mur-Labadia et al. \cite{mur2023multi} build a 3D multi-label mapping of affordances, which was exploited in task-oriented path planning.

In terms of perception affordance models, supervised approaches learn by direct supervision of manual annotated affordances masks, providing pixel-wise precision and more accurate location, very beneficial for robot grasping. The seminal UMD dataset \cite{myers2015affordance} provides mask annotations for RGB-D images, but the dataset is composed of pre-defined objects captured in isolated conditions. The IIT-Aff dataset \cite{nguyen2017object} comprises real-world objects in multiple environments, enabling a better generalization in real conditions. Synthetic data reduces the cost of manual annotations, but the simplicity of the supervised synthetic images and the dataset gap limits its performance in real-world scenarios. Nguyen et al. \cite{nguyen2016detecting} introduced a convolutional encoder-decoder to learn affordances from a deep latent space, which was further refined using feature maps with Conditional Random Fields in a post-processing step in a subsequent work \cite{nguyen2017object}.  Affordance-Net \cite{do2018affordancenet} adapted an object detection for affordance perception with three significant contributions: a multi-task loss function, a resizing strategy, and a sequence of deconvolutional layers for producing high-resolution masks. 
Minh et al. \cite{minh2020learning} integrated ResNet101 for superior feature extraction and the Feature Pyramid Networks (FPN) to amalgamate multi-scale features, thereby elevating segmentation precision. 
Caselles et al. \cite{caselles2021standard} demonstrated the effectiveness of reusing standard instance segmentation models for affordance perception. 
\revision{Recently, Apicella et al.~\cite{apicella2024segmenting} introduced M2F-AFF an affordance segmentation model based on the popular Mask2Former architecture.}
Weakly supervised approaches learn from watching human-object interactions \cite{fang2018demo2vec, nagarajan2020ego} but obtain diffuse interaction hotspots rather than precise affordance masks on the object part. 
Demo2Vec \cite{fang2018demo2vec} infer affordance key points on a static object by watching demonstration videos of the respective interaction using only the action label as supervision. Similarly, Nagarajan et al. \cite{nagarajan2019grounded} obtained interaction hotspots by extracting gradient-weighted attention maps from training an action classifier on videos. 
\textcolor{black}{However, any previous work extracted uncertainty from visual affordances segmentation, reducing the interpretability and robustness of model's prediction.}

\revision{Recent advances in visual affordance learning have explored beyond supervised approaches. Instead of relying on predefined labels, affordance grounding methods focus on learning precise, actionable affordances from egocentric videos. Li et al.~\cite{li2025learning} propose a framework for learning precise manipulation affordances from egocentric videos, leveraging geometric reasoning and foundation models to extract contact-rich affordance maps that transfer effectively to robotic grasping tasks. Extending to complex interactions, Heidinger et al. introduce 2HandedAfforder~\cite{heidinger20252handedafforder}, a dataset and learning framework for capturing fine-grained bimanual affordances from human demonstration videos, enabling robots to infer coordinated two-handed manipulation strategies.
A significant trend involves leveraging Large Language Models (LLMs) and Vision-Language Models (VLMs) to ground affordances in open-vocabulary contexts \cite{qian2024affordancellm} and analyze the zero-shot capabilities of models like CLIP \cite{cuttano2024does}. Similarly, Li et al. introduce One-Shot Open Affordance Learning (OOAL)~\cite{li2024one}, a framework that learns from a single annotated example per base class and transfers affordance knowledge to novel categories via text–vision alignment. Moving beyond passive grounding, Tang et al. propose CoTDet~\cite{Tang_2023_ICCV}, which integrates affordance knowledge prompting into task-driven object detection using chain-of-thought reasoning to enhance task-aware perception. Furthermore, the domain is expanding into 3D environments and reasoning-heavy tasks. This is evidenced by the development of RAGNet for reasoning-based segmentation \cite{wu2025ragnet}, AffordBot for 3D fine-grained embodied reasoning \cite{wangaffordbot}, and GEAL for cross-modal 3D consistency \cite{lu2025geal}. To reduce reliance on dense annotations, Moon et al.~\cite{moon2025selective} introduce a selective contrastive learning framework for weakly supervised affordance grounding, exploiting cross-view consistency and discriminative feature alignment to localize affordances without pixel-level supervision.

A comprehensive overview of the field is provided by Apicella \emph{et al.}~\cite{apicella2025visual}, who analyze trends, benchmarks, and reproducibility challenges, highlighting the sensitivity of affordance segmentation models to experimental settings. 
}

\subsection{Neural Network Uncertainty Quantification}

\revision{The shift toward large-scale AI has reignited the discussion on the necessity of robust uncertainty quantification, with recent position papers arguing that Bayesian Deep Learning and Uncertainty Quantification are essential for the reliability and safety of modern foundational models \cite{papamarkou2024position}.}

Sampling-based methods \cite{gal2016dropout, durasov2021masksembles} for \revision{Bayesian deep learning} incorporate intermediate sampling layers to train a single model, quantifying the uncertainty by comparing the divergence in the different predictions. Similarly, ensemble methods \cite{huang2017snapshot, lakshminarayanan2017simple} require training multiple models to capture the diversity in the predictions. \revision{Recently, Wild et al.~\cite{wild2023rigorous} establish a rigorous connection between deep ensembles and variational Bayesian methods, showing that ensembles can be interpreted as approximate posterior inference under specific variational objectives. BELLA~\cite{doan2025bayesian} is a variational Bayesian neural network method that approximates the weight posterior with a low-rank covariance Gaussian that captures correlations between parameters, providing a principled Bayesian analogue to deep ensembles, which can be interpreted as implicitly sampling from multiple posterior modes—an interpretation theoretically grounded by the variational perspective on ensembles in~\cite{wild2023rigorous}.}

Feature-space techniques \cite{postels2020quantifying} estimate the uncertainty with a single-pass by measuring the distance or density of the sample compared with the training data distribution in the new space. 
Attention-based architectures have been also extended for uncertainty quantification. 
BayesFormer \cite{sankararaman2022bayesformer} applies a variational inference-based dropout framework within the transformer architecture, demonstrating its efficacy in NLP. \revision{For certain applications, transformers are able to provide uncertainty quantification directly. Müller et al.~\cite{mullertransformers} demonstrate that transformers trained on synthetic tasks can learn to approximate Bayesian posterior inference in context, effectively implementing amortized inference through attention mechanisms. Their results show that, given appropriate training distributions, transformers can reproduce classical Bayesian updating behavior without explicitly encoding probabilistic structure. Reuter et al.~\cite{reuter2025can} analyze in-context learning where transformers can learn full Bayesian inference and, under certain contitions, are able to  approximate the posterior predictive distributions faithfully.}
On active learning tasks, Gleave et al. \cite{gleave2022uncertainty} propose a last-layer ensemble for uncertainty estimation. However, their estimated epistemic uncertainty is poorly calibrated since the ensemble models are very similar to each other. 

The uncertainty estimation increases the interpretability and robustness of deep neural networks. \revision{Recently, Wang and Ji~\cite{wang2024beyond} examine the relationship between Bayesian neural networks and evidential deep learning, moving beyond Dirichlet-based uncertainty models. This work bridges two prominent paradigms in uncertainty modeling, offering insights into how subjective evidence representations connect to Bayesian principles.}
In terms of perception, as Kendall et al. \cite{Gal} shown on computer vision applications, the epistemic term appears in challenging pixels or occluded objects out of the distribution. 
Oppositely, aleatoric uncertainty appears in the object's contours or in far-away regions with higher camera noise. 
Ji et al. \cite{ji2022fast} introduced a re-calibration strategy to improve the accuracy of camouflaged object detection.
Morilla et al. \cite{morilla2023robust} combine Bayesian semantic predictions to mitigate the effect of overconfident outlier predictions and build more robust semantic maps. 
Kim et al. \cite{kim2023uncertainty} proposed a semi-supervised approach to few-shot segmentation that enhances performance by incorporating unlabeled images through uncertainty-guided pseudo label refinement.
\textcolor{black}{However, prior work has not extended instance segmentation models to fully Bayesian formulations, nor provided a systematic comparison of sampling-based approaches for uncertainty estimation.} 



\section{Bayesian Instance Segmentation of Affordances}


\textcolor{black}{Prior works \cite{caselles2021standard, do2018affordancenet} consider affordances segmentation as a deterministic problem and do not evaluate uncertainty quantification. 
Therefore, our contribution in this letter is a novel Bayesian Instance Segmentation of affordances, which segments objects-part affordances and quantifies both the spatial uncertainty related to the predicted masks and the semantic uncertainty, associated with the affordance class of each detection. We also contribute with an extensive comparative of different sampling-based techniques and the incorporation of an attention-based backbone in the detector.}

\textcolor{black}{We first introduce the base architecture, Mask R-CNN, for object affordance segmentation (Section~\ref{sec: maskrcnn}). We then review the theoretical foundations of Bayesian Deep Learning, which we adopt for uncertainty estimation (Section~\ref{sec: uncer_fundam}). Finally, we describe how we integrate these principles into the instance segmentation model (Section~\ref{sec: join_both}). Our resulting approach performs object affordance segmentation while estimating both spatial uncertainty—associated with object mask boundaries—and semantic uncertainty—related to affordance classes. Moreover, our method disentangles the contributions of epistemic and aleatoric uncertainty.}

\subsection{Mask-RCNN for Instance segmentation of affordances}
\label{sec: maskrcnn}

\begin{figure}[t]
\centering
\includegraphics[width=0.99\columnwidth]{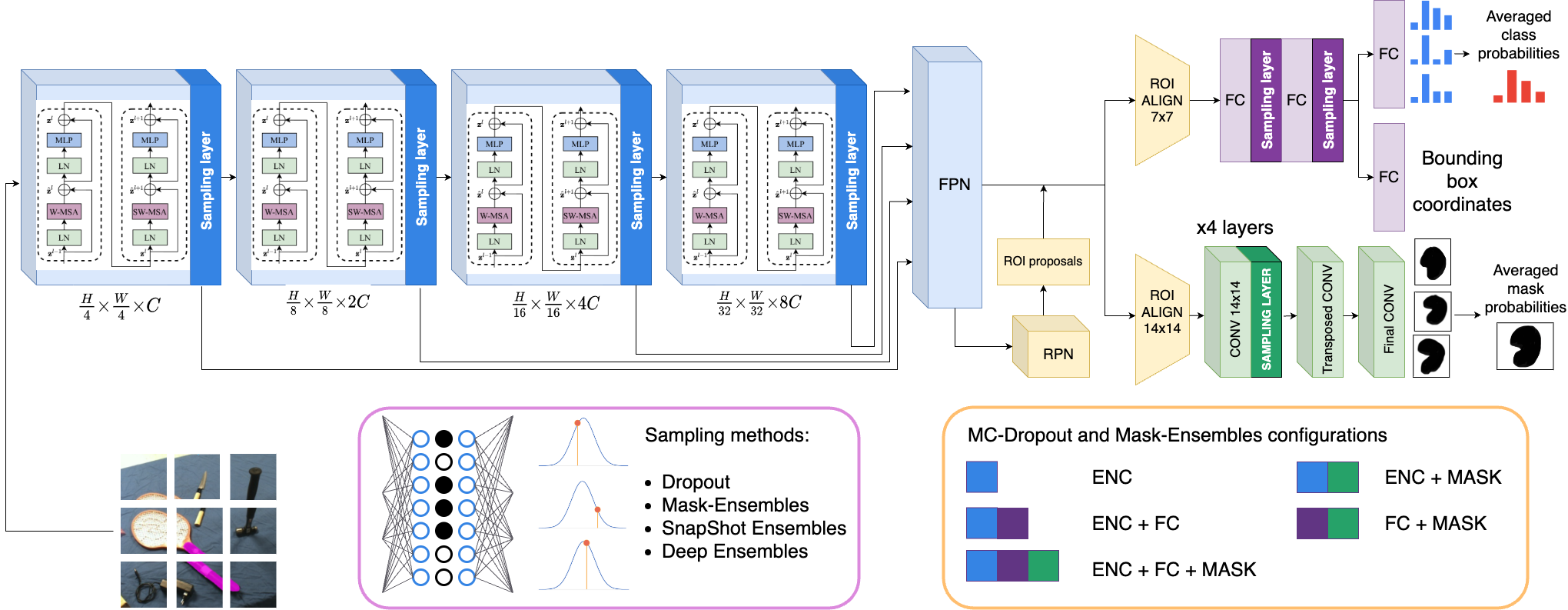}
\caption{Our model architecture is composed of an attention-based backbone \cite{liu2021swin} extended with sampling layers \cite{gal2016dropout, durasov2021masksembles}. We compare the performance and calibration of different sample-based (MC-Dropout \cite{gal2016dropout}, Mask-Ens \cite{durasov2021masksembles}) and ensembles (Deep-Ens \cite{lakshminarayanan2017simple}, Snap-Ens \cite{huang2017snapshot}) approaches for the Bayesian inference.}
\label{fig:maskrcnn}
\end{figure}

Mask R-CNN is a two-stage instance segmentation framework that extends Faster R-CNN by jointly predicting object bounding boxes, class labels, and pixel-wise instance masks.
Figure \ref{fig:maskrcnn} illustrates the general architecture of our approach.
First, an encoder extracts hierarchical multi-scale feature maps from the input image. 
\textcolor{black}{Previous works~\cite{aff_loren, caselles2021standard, ren2015faster} typically rely on a ResNet-50 backbone, a convolutional architecture with limited representational capacity. 
In this work, we instead adopt the Swin Transformer~\cite{liu2021swin} as the backbone of Mask R-CNN. This architecture processes images via hierarchical self-attention, progressively merging local image patches while reducing spatial resolution and increasing feature dimensionality, thereby enabling richer and more scalable representations.}
This structure mirrors the pyramid-like design of convolutional-based encoders, enabling efficient processing of high-resolution images and the extraction of multi-scale feature maps.
The encoder progresses through multiple stages, each consisting of several Swin Transformer blocks that process the input image at different scales. Each block includes a Window-based Multi-head Self-Attention (W-MSA) module, a Multi-Layer Perceptron (MLP), a Shifted-Window Multi-head Self-Attention (SW-MSA) module, and a second MLP.
W-MSA computes self-attention within non overlapping windows, capturing local information, while SW-MSA shifts these windows to share patches with the previous layer, facilitating effective global information integration while maintaining linear complexity with the image size. 
LayerNorm (LN) layers and residual connections are applied around each module to enhance stability and convergence.
Additionally, we integrate a Feature Pyramid Network (FPN) 
with lateral connections and top-down propagation, which produces high-resolution feature maps at all levels, enriched with high-level semantic features.

Based on the FPN features, the Region Proposal Network (RPN) generates Region Proposals (RPs) that highlight areas likely to contain objects.
The Region of Interest (RoI) Align 7 $\times$ 7  block then standardizes the size of the proposed RPs, ensuring that the extracted features correspond to the input regions. 
Subsequently, we obtain the semantic class and the bounding box coordinates of each RPs using respective Multi-Layer Perceptrons (MLPs), containing intermediate sampling layers.
Then, a softmax converts the affordance class logit predictions into a probability vector $\mathbf{p}^{c}$. 
In parallel, we obtain the mask prediction after applying a 14 $\times$ 14 RoI Align block, four convolutional layers with intermediate sampling layers, and a transposed convolution to upsample the mask resolution. The binary probability masks ${\mathbf{p}}^{h}$ are obtained after a final convolution layer and a sigmoid.

We train the model end-to-end following \cite{he2017mask} with a multi-task loss applied on each RoI as $\mathcal{L} = \mathcal{L}_{cls} + \mathcal{L}_{box} + \mathcal{L}_{mask} + \mathcal{L}_{RPNbox} + \mathcal{L}_{RPN cls} \label{eq:loss}$. The classification loss $\mathcal{L}_{cls}$ computes the log loss to classify each anchor as an object or background accurately. The bounding box regression $\mathcal{L}_{box}$ refines the coordinates of the RP bounding boxes based on the ground truth offsets with a smooth L1 loss. The mask loss $\mathcal{L}_{mask}$ computes only for the positive instances the binary cross-entropy loss at the pixel level for the mask prediction. The RPN classification loss $\mathcal{L}_{RPN cls}$ computes a cross-entropy loss to train the RPN to distinguish between foreground and background anchors, while the RPN bounding box loss $\mathcal{L}_{RPN box}$ applies a smooth L1 regression to refine the anchor box predictions generated by the RPN.
\textcolor{black}{However, Mask-RCNN only produces \emph{deterministic detections}, ignoring modeling prediction uncertainty, which limits the robustness and interpretability of the predictions. To this end, we extend this approach with Bayesian deep learning techniques.}

\subsection{Uncertainty estimation in Bayesian deep learning}
\label{sec: uncer_fundam}

\textcolor{black}{Uncertainty estimation requires reasoning beyond point predictions and modeling the full distribution of possible model outputs. In Bayesian deep learning, this is achieved by propagating uncertainty in the model parameters $\mathbf{w}$ to the prediction space through the \emph{predictive posterior distribution},}

\begin{equation}
   p(y^* \lvert X^*, \mathcal{D}) = \int_\mathcal{W} p(y^* \lvert X^*, \mathbf{w})  p(\mathbf{w} \lvert \mathcal{D}) d\mathbf{w},
   \label{eq:predpost}
\end{equation}
\textcolor{black}{where $\mathcal{D}=\{\mathbf{X},\mathbf{Y}\}$ denotes the training dataset of $N$ input–label pairs $\{(\mathbf{X}_i,\mathbf{y}_i)\}_{i=1}^N$, and $y^*$ is the prediction for a new input $\mathbf{X}^*$. Computing the predictive posterior requires evaluating the full posterior distribution over the model parameters,}


\begin{equation}
     p(\mathbf{w} \lvert \mathcal{D}) = \frac{p(\mathcal{D} \lvert \mathbf{w}) p(\mathbf{w})}{p(\mathcal{D})} = \frac{p(\mathcal{D} \lvert \mathbf{w}) p(\mathbf{w})}{\int_\mathcal{W} p(\mathcal{D} \lvert \mathbf{w}) p(\mathbf{w}) d \mathbf{w}}
\end{equation}

where the marginal likelihood $p(\mathcal{D})$ involves another high-dimensional integral,
\textcolor{black}{In deep neural networks, this integral is intractable, making exact Bayesian inference impractical. To address this challenge, sampling-based approaches~\cite{gal2016dropout} approximate the true posterior $p(\mathbf{w} \mid \mathcal{D})$ with a finite set of sampled network weights $\{\mathbf{w}_m\}_{m=1}^M$. The predictive posterior in Equation~\ref{eq:predpost} is then approximated via Monte Carlo integration as}
\begin{equation}
     \hat{p}(y^* \lvert X^*, \mathcal{D}) \approx \frac{1}{M} \sum_{m = 1}^M f(\mathbf{X}^*, \mathbf{w}_m),  
     \label{eq:predpostmc}
\end{equation}
\textcolor{black}{where $f(\cdot,\mathbf{w}_m)$ denotes the network output for the $m$-th weight sample. In practice, each forward pass produces a class probability vector $\mathbf{p}_m = f(\mathbf{X}^*, \mathbf{w}_m)$, and the predictive distribution is estimated as their empirical mean,}
\begin{equation}
     \hat{p}(y^* \lvert X^*, \mathcal{D}) \approx \frac{1}{M} \sum_{m = 1}^M \mathbf{p}_m. 
     \label{eq:predpostmc}
\end{equation}

\textcolor{black}{In this letter, we compare several methods for approximating the true posterior distribution $p(\mathbf{w} \lvert \mathcal{D})$, each relying on a different strategy to sample the neural network weights $\{\mathbf{w}_m\}_{m=1}^M$. As we show in Figure \ref{fig:maskrcnn}, we incorporate sampling layers at different stages of the neural network in order to generate probabilistic latent embeddings.}


\begin{enumerate}[label=\alph*)]
    \item Monte Carlo (MC)-Dropout, introduced by Gal et al. \cite{gal2016dropout}, obtains $M$ weight samples by first training the network and obtaining an optimal set of parameters from the maximum posterior distribution $\mathbf{w}^* = \mathbf{w}_{MAP}$. The sampling distribution is then generated by applying dropout at inference, effectively multiplying the optimal weights by a Bernoulli-distributed mask, $\mathbf{Ber}(d)$, which determines whether a neuron is dropped. This results in the sampled weights $\{ \mathbf{w}_m \}_{m=1}^{M}$, where $\mathbf{w}_m = \mathbf{w}^* \cdot z_m$ with $z_m \sim \mathbf{Ber}(d)$.
    In practice, this is equivalent to keeping dropout layers active at inference. 
    Although MC-Dropout effectively minimizes the Kullback-Leibler divergence between the true posterior $p (\mathbf{w} \lvert \mathcal{D})$ and a tractable variational approximation $q_\theta (\mathbf{w})$ \cite{gal2016dropout}, the correlation between $\mathbf{w}_m$ samples may lead to an underestimation of the uncertainty. 
    Additionally, the use of dropout results in predictions based on effectively smaller networks due to dropped neurons, which can degrade performance.

    \item Mask-Ens \cite{durasov2021masksembles} also trains a single model with intermediate sampling layers, similar to MC-Dropout. However, unlike the stochastic nature of random dropout, Mask-Ens pre-computes a set of binary masks that, when applied, create $M$ sampling layers with mask-based dropout.
    Mask-Ens offer two key advantages over MC-Dropout. First, the sampling masks exhibit lower correlation, leading to improved calibration performance. Second, because the masks can be predefined, the layer size can be adjusted to compensate for dropped neurons, ensuring a constant number of active neurons equivalent to the original network without dropout.
    The mask generation process incorporates a controllable degree of overlap, regulated by a scale parameter $s$. At inference, the individual model weights $\{\mathbf{w}_m \}_{m=1}^{M}$ are obtained as $\mathbf{w}_m = \mathbf{w}^* \cdot \text{Mask}(s, m)$.

    \item Deep-Ens \cite{lakshminarayanan2017simple} trains an ensemble of $M$ models with randomly initialized network parameters, dataset shuffling, and random data augmentation. 
    This method ensures that the weights $\mathbf{w}_m$ converge to different local optima. Notably, it effectively captures the multi-modality of the posterior distribution $p(\mathbf{w} \lvert \mathcal{D})$, even with a relatively small number of ensemble \cite{lakshminarayanan2017simple}.

    \item Snap-Ens \cite{huang2017snapshot} mitigates the computational overhead of Deep-Ens, which requires training $M$ models from scratch, by instead training all $M$ models sequentially within a single training loop. This is achieved by cyclically increasing the learning rate once a model has converged, leading to $M$ distinct solutions that capture multiple local minima throughout training. Snap-Ens does not require architectural modifications, as $\mathbf{w}_m$ are extracted at different training stages, making it easily applicable to pre-trained models.
\end{enumerate}




\subsection{Bayesian Instance Segmentation}
\label{sec: join_both}
 
\textcolor{black}{In this section, we describe how we extend a classical deterministic object detector, such as Mask R-CNN, into a Bayesian detector capable of segmenting object affordances while providing well-calibrated uncertainty estimates. 
During training, we introduce intermediate stochastic sampling layers while preserving the original training procedure \cite{he2017mask} and loss functions defined in Equation \ref{eq:loss}. At inference time, as the multiple stochastic forward passes of the sampling techniques produce a set of detections, we define a merging strategy and derive uncertainty estimates from the resulting predictive distribution.}

\begin{enumerate}
    \item \textbf{Sampling.} We conduct $M$ forward passes to generate multiple detections per inference pass. Each detection $\mathbf{d} = \{\mathbf{b}, \mathbf{p}^{c}, \mathbf{p}^{h} \}$ consists of a bounding box $\mathbf{b} =(x,y,w,h)$, a class probability vector $\mathbf{p}^{c}$ and a \emph{heatmap} representing the binary probability of each pixel belonging to the detection or background $\mathbf{p}^{h}$. Depending on the uncertainty extraction method (Deep-Ens, Snap-Ens, MC-Dropout, Mask-Ens), the modification of the network weights $\mathbf{w}_m$ differs in each of the $M$ forward passes. 

    \item \textbf{Clustering.} As defined by Miller et al. \cite{miller2019evaluating}, an observation is a collection of detections with high spatial and semantic affinity $\mathcal{O}_d= \{\mathbf{d}_1,...,\mathbf{d}_k\}$. Note that the number of times an instance is detected, $k$, is always less than or equal to the number of forward passes, $M$, i.e., $k \leq M$.
    Following \cite{miller2019evaluating}, we cluster detections into observations using the Basic Sequential Algorithm Scheme (BSAS) with the Intersection over Union (IoU) of the detected masks $\mathbf{p}^{h}$ as the similarity metric. 
    \textcolor{black}{All detections from the $M$ forward passes are pooled together prior to clustering, allowing the algorithm to handle cases where different passes yield different numbers of detections.}
    We only group detections with the same label class to avoid the merging of overlapped objects whose masks are spatially close.
    \textcolor{black}{This design choice follows~\cite{miller2019evaluating}, which shows that enforcing class consistency during clustering yields higher detection accuracy and prevents incorrect merging of distinct objects, while class uncertainty is later captured through the predictive distribution.}
    Compared to previous approaches limited to object detection with bounding boxes \cite{miller2019evaluating}, we compare pixel-wise masks of instances, enabling improved merging of clustered or irregularly shaped objects. 

    \item \textbf{Averaging.} Following the Bayesian approach in Equation \eqref{eq:predpostmc}, we average the clustered detections to obtain the final predictions $\mathcal{O} = \{ \bar{\mathbf{b}}, \mathbf{\bar{p}}^{c}, \mathbf{\bar{p}}^{h} \}$. 
    Consequently, an observation consists of a set of detections from different sampling steps, representing an approximation of the true posterior distribution of the observation.
    
    \begin{equation}
        \mathcal{O} = \Big\{ \frac{1}{M} \sum_{m=1}^M \mathbf{b}_m, \frac{1}{M} \sum_{m=1}^M \mathbf{p}^{c}_m, \frac{1}{M} \sum_{m=1}^M \mathbf{p}^{h}_m \Big\} 
    \end{equation}

    \item \textbf{Uncertainty Estimation.} Our novel Bayesian instance segmentation approach extends deterministic models by incorporating per-pixel uncertainty estimation, $\sigma$, which accounts for both epistemic and aleatoric contributions.
    \begin{equation}
        \sigma = \sigma_e + \sigma_a
    \end{equation}
    
    The epistemic uncertainty $\sigma_e$ represents the variability of the model parameters $\mathbf{w}$, which can be used to identify when a sample is out of the training distribution. Consequently, it can be reduced with a larger and more diverse dataset, as the model incorporates more knowledge.
    Following \cite{Gal}, we compute the epistemic covariance matrix as the average difference between the squared matrix formed by multiplying $\mathbf{p}_m - \bar{\mathbf{p}}$ by its transpose, as shown in Equation \eqref{eq:epistemic}.
    It quantifies the deviation of each prediction $\mathbf{p}_m$ from the mean probability $\bar{\mathbf{p}}$, which is computed as the average of the sampled probability vectors $\bar{\mathbf{p}} = \frac{1}{M} \sum_{m=1}^M \mathbf{p}_m$.
    
    \begin{equation}
         \sigma_{e} = \frac{1}{M} \sum_{m=1}^{M} \left(\mathbf{p}_m - \bar{\mathbf{p}}\right)  \cdot \left(\mathbf{p}_m - \bar{\mathbf{p}}\right)^T
         \label{eq:epistemic}
    \end{equation}
    
    The aleatoric uncertainty $\sigma_a$ is inherent to data noise and cannot be reduced with more data, reflecting observations noise like the camera motion or object boundaries. Since it is intrinsic to the data distribution, collecting more data does not reduce it. The aleatoric covariance matrix is the average difference between a diagonal matrix formed with the components of the $m$-th prediction vector $\mathbf{p}_m$ and the outer product of the prediction vector $\mathbf{p}_m$ with itself.

    \begin{equation}
         \sigma_{a} = \frac{1}{M} \sum_{m=1}^{M} \mathbf{diag}(\mathbf{p}_m) - \mathbf{p}_m \cdot \mathbf{p}_m^T
    \end{equation}
    
    Previous works \cite{miller2019evaluating, aff_loren} either ignore the differences between the two sources of uncertainty or solely evaluate the variance associated with mask segmentation. We introduce a novel formulation for extracting both spatial uncertainty $\sigma^{sp}$ and semantic uncertainty $\sigma^{sem}$, as well as their respective epistemic and aleatoric variance contributions.
    
    Spatial uncertainty is derived from the binary probability mask vector $\mathbf{p}^{h}$ and captures pixel-level differences in  the segmentation.
    Analytically, we compute
    $\sigma^{sp} = \sigma_a^{sp} + \sigma_e^{sp}$,
    where $\mathbf{p}_m  = \mathbf{p}_m^{h}$ represents the output after the sigmoid of the final mask convolution in the $m$-th forward pass and therefore, the spatial uncertainty $\sigma^{sp}$ is extracted per-pixel and distributed in a 2D map along the image.
    
    Semantic uncertainty reflects the ambiguities associated with the semantic affordance class and it is constant along all the detected object pixels. 
    Similarly, we obtain $\sigma^{sem} = \sigma_{a}^{sem} + \sigma_{e}^{sem}$ where $\mathbf{p}_m = \mathbf{p}_m^{c}$ represents the probability of the affordance class, computed as the softmax output associated with the prediction of the semantic class of the $m$-th sampling. Therefore, we assume that the semantic uncertainty $\sigma^{sem}$ is constant for all the pixels within the predicted affordance mask. We collapse the respective covariance by summing the trace of the matrix.
\end{enumerate}

\section{Experiments}

Following previous works \cite{caselles2021standard, do2018affordancenet}, we adopt the $F_\beta^w$ score \cite{margolin2014evaluate} to evaluate the affordance segmentation performance. This score assesses foreground maps and adjusts pixel-level errors based on three key assumptions: dependency, interpolation, and equal importance.
Next, we assess the uncertainty of probabilistic affordance detections using our novel Probabilistic Mask Quality (PMQ) metric. Additionally, we evaluate the calibration of the predictions through Expected Calibration Error (ECE) and Area Under the Sparsification Curve (AUSE).
We provide a detailed calibration analysis for both spatial uncertainty (associated with mask contours) and semantic uncertainty (associated with the affordance class).





\subsection{Probabilistic Mask Quality metric}

We design our novel PMQ inspired by the Probabilistic-based Detection Quality (PDQ) score \cite{hall2020probabilistic}. While the PDQ analyzes the spatial and semantic uncertainties of probabilistic object detections based on bounding box predictions, our novel PMQ score compares the averaged predicted masks $\bar{\mathbf{p}}^{h}$, providing a more fine-grained evaluation and preserving boundary detections. 
Compared to metrics based on the Average Precision score, PMQ does not rely on arbitrary IoU thresholds that filter the number of detections, nor does it evaluate foreground/background quality separately. 


We define each $i$ ground-truth object annotation as $\mathcal{G}_i = \{\hat{\mathbf{b}}_i, \hat{\mathbf{c}}_i, \hat{\mathcal{M}}_i \}$, where $\hat{\mathbf{b}}_i$ is the ground-truth bounding-box coordinates, $\hat{c}_i$ the discrete label class and $\hat{\mathcal{M}}_i$ the mask around the object. The output of our probabilistic model is a set of $j$ averaged detections, where $\mathcal{O}_j = \{\bar{\mathbf{b}}_j, \bar{\mathbf{p}}^{c}_j, \bar{\mathbf{p}}^{h}_j \}$. We start evaluating the semantic estimation with the label quality $Q_l$. It is the probability score of the respective ground-truth class and not the highest predicted class score.

\begin{equation}
    Q_L (\mathcal{G}_i, \mathcal{O}_j) =[\mathbbm{1}_{c= \hat{c}_i}]^T \cdot \bar{\mathbf{p}}^{c}_j 
\end{equation}

where $\mathbbm{1}_{c= \hat{c}}$ is the indicator function, being 1 for $c= \hat{c}$ and 0 otherwise.
The spatial quality $Q_s$ measures how the probabilistic mask captures the contour of the ground-truth affordance mask. It is the exponential of the negative sum between the foreground loss $L_{FG}$ and the background loss $L_{BG}$ at the mask level.

\begin{equation}
    Q_S(\mathcal{G}_i, \mathcal{O}_j) = \textrm{exp} (-(L_{FG}(\mathcal{G}_i, \mathcal{O}_j ) + L_{BG}(\mathcal{G}_i, \mathcal{O}_j )))
\end{equation}

While the computation of the foreground loss $L_{FG}$ and the background loss $L_{BG}$ is similar, they evaluate opposite effects. The $L_{FG}$ penalizes predicted pixels inside the ground truth mask with low probability, while the $L_{BG}$ penalizes pixels outside the ground truth mask with high probability. The $L_{FQ}$ is the average negative logarithm of the mask probabilities $\bar{\mathbf{p}}^{h}_j$ inside the ground-truth segments. The $L_{BG}$ is defined as the sum of the negative log-pixel probabilities evaluated in the set $\mathcal{V}_{ij} = \{ \bar{\mathbf{p}}^{h}_j -  \hat{\mathcal{M}}_i\}$, corresponding to detected pixels $\bar{\mathbf{p}}^{h}_j$ but outside the ground truth masks $\hat{\mathcal{M}}_i$.

\begin{equation}
    L_{FG}(\mathcal{G}_i, \mathcal{O}_j ) = - \frac{1}{\lvert  \hat{\mathcal{M}}_i \lvert} \sum_{x \in  \hat{\mathcal{M}}_i} \textrm{log} (\bar{\mathbf{p}}^{h}_j(\text{x})))
\end{equation}

\begin{equation}
    L_{BG}(\mathcal{G}_i, \mathcal{O}_j ) = - \frac{1}{\lvert \hat{\mathcal{M}}_i \lvert} \sum_{x \in \mathcal{V}_{ij}} \textrm{log} (1 - \bar{\mathbf{p}}^{h}_j(\text{x}))
\end{equation}

Then, we compute the pairwise PMQ (pPMQ) for each observation $\mathcal{O}_j$ and object label $\mathcal{G}_i$ as the geometric mean of the respective semantic $Q_L$ and spatial $Q_S$ qualities.

\begin{equation}
    pPMQ(\mathcal{G}_i, \mathcal{O}_j) = \sqrt{Q_S (\mathcal{G}_i, \mathcal{O}_j ) \cdot Q_L(\mathcal{G}_i, \mathcal{O}_j )}
\end{equation}
At this point, the pairwise PMQ contains a PMQ value between each label $G_i$ and observation $\mathcal{O}_j$ pair.  We apply the Hungarian algorithm \cite{kuhn1955hungarian} to obtain the optimal assignment of the $N_{TP}$ true positives, denoted as $\mathbf{q} = [q_1, ..., q_{TP} ]$, between the observations $\mathcal{O}_j$ and the label $\mathcal{G}_i$. An optimal assignment is zero when a ground-truth object is not detected (false negative) or a observation does not represent any affordable object-part (false positive). The final PMQ, compute along all the frames $N_F$ in the evaluation set is computed by averaging the $pPMQ$ along the total number of false positives $N_{FP}$ (the observation does not match with the ground-truth), the total false negatives $N_{FN}$ (no exits ground-truth) and all the true positives $N_{TP}$.


\begin{equation}
    PMQ(\mathcal{G}, \mathbf{d}) = \frac{1}{\sum_{f = 1}^{N_F} \left(N_{TP}^f + N_{FN}^f + N_{FP}^f\right)} \sum_{f = 1}^{N_F} \sum_{i = 1}^{N_{TP}} \mathbf{q}^f(i)
\end{equation}


\subsection{Calibration metrics}

We evaluate model calibration with the Expected Calibration Error (ECE) and the Area Under the Sparsification Error (AUSE). The ECE \cite{guo2017calibration} measures the absolute calibration error of a probabilistic classification. A well-calibrated classification model produces predictions where the predicted confidence accurately reflects the likelihood of being correct.
To compute ECE, predictions are first grouped into $N$ bins based on their confidence scores. The calibration error is then measured as the difference between the accuracy and confidence of each bin. The bin accuracy, $\text{acc}(B_n)$, is the fraction of correctly classified instances within the bin. The bin confidence, $\textrm{conf}(B_n)$, is the average predicted probability of the instances in the bin:

\begin{equation}
    \textrm{ECE} = \sum^N_{n = 1} \frac{\lvert B_n \lvert}{N_s} \lvert \textrm{acc}(B_n) - \textrm{conf}(B_n) \lvert
\end{equation}

where $N_s$ is the total number of samples, and $B_n$ is the number of samples in the respective bin.

The AUSE metric provides a relative measure of uncertainty, computed as the difference between the estimated uncertainty and the true error of the model, measured in terms of the Brier Score (BS) \cite{brier}. The BS is the mean squared difference between the predicted probabilities $\text{p}$ and the corresponding one-hot ground-truth vector $\mathbbm{1}_{c= \hat{c}}$, being 1 for $c= \hat{c}$ and 0 otherwise:

\begin{equation}
    \textrm{BS} = \frac{1}{N_p} \sum_{n = 1}^{N_p} (\text{p}_n - \mathbbm{1}_{c= \hat{c}})^2
    \label{eq:brier}
\end{equation}

On the plot, we gradually remove pixels based on their true error (Oracle) or their estimated variance (Model). We then normalize and compute the area between these curves, reporting the sparsification error (AUSE). The lower this value, the better the estimated variance represents the true error.
In Equation \eqref{eq:brier}, for the semantic AUSE, $N_p$ is the total number of detections in the entire test dataset, while for the spatial AUSE, $N_p$ is the total number of pixels with an estimated variance (thus, background regions between bounding boxes are ignored).

\subsection{Implementation details}

We train the models with ResNet-50 and ResNeXt-101 using the Adam optimizer with learning rates of $10^{-3}$ and $10^{-5}$, respectively.
The models with Swin-T \cite{liu2021swin} as the encoder were trained with AdamW, a learning rate of $10^{-5}$, a weight decay of 0.05, and $\beta = (0.9, 0.99)$.
We initialize the weights with the respective pre-trained versions on COCO to ensure convergence and improve performance, starting with a linear warm-up of the learning rate during the first 1,000 iterations.
We run our experiments in an NVIDIA GeForce 4090 GPU.
We conducted the experiments on the IIT-AFF dataset \cite{nguyen2017object}, which consists of 8,835 real-world images depicting seven different affordance categories (contain, cut, display, engine, grasp, hit, pound, support, and w-grasp). The dataset is specifically designed for robotics-based scenarios and represents common manipulation capabilities for autonomous agents. We also report results in the UMD dataset \cite{myers2015affordance}, which contains 28,843 images of 17 object categories and 7 affordance actions, captured on a rotating table in clutter-free conditions.

\begin{table}[t!]
    \centering
    \begin{tabular}{l|cccc|c}
    \multicolumn{1}{c}{} & \multicolumn{4}{c}{Backbone} & \\ 
    \cmidrule(lr){2-5}
    Method & VGG & R50 & R101 & Swin-T & $F_\beta^w$ \\
    \midrule
     ED-RGB \cite{nguyen2016detecting}& \checkmark& -& -& -&57.64\\
     R-FCN \cite{nguyen2017object}& -& -& \checkmark& -&69.62\\    
    AffordanceNet \cite{do2018affordancenet} & \checkmark &  -&  -&  -& 79.90 \\ 
    BPN \cite{yin2022object} &  -& \checkmark &  -&  -& 79.64\\ 
    GSE \cite{zhang2022multi} &  -&  -& \checkmark &  -& 82.33\\ 
    Mask-RCNN \cite{caselles2021standard} &  -& \checkmark &  -&  -& 84.40\\
 \midrule  
    Ours, Bayesian &  -&  \checkmark &  -& -& 86.90\\ 
    Ours, Deterministic&  -&  -&  -& \checkmark & 88.32\\ 
    Ours, Bayesian&  -&  -&  -& \checkmark & \textbf{90.60}\\ 
    \bottomrule
    \end{tabular}
    \caption{Affordance segmentation comparative with the state-of-the-art in the IIT-Aff dataset.}
    \label{tab:IIT_Aff_sota}
\end{table}

\begin{table}[t!]
    \centering
    \begin{tabular}{l|cccc|c}
    \multicolumn{1}{c}{} &  \multicolumn{4}{c}{Backbone}& \\ 
    \cmidrule(lr){2-5}
    Method & VGG &R50 & R101 & Swin-T & $F_\beta^w$ \\ 
    \midrule
    AffordanceNet \cite{do2018affordancenet} & \checkmark &-&  -&  -& 79.90 \\ 
    BPN \cite{yin2022object} &  -&\checkmark &  -&  -& 86.21\\ 
    GSE \cite{zhang2022multi} &  -&-& \checkmark &  -& 85.50 \\ 
    Mask-RCNN \cite{caselles2021standard} &  -&\checkmark &  -&  -& 84.21 \\
        \revision{M2F-AFF \cite{apicella2024segmenting}} &  -&\revision{\checkmark} &  -&  -& \revision{79.93} \\
    \midrule
    Ours, Deterministic&  -&-&  -& \checkmark & 86.03\\ 
    Ours, Bayesian&  -&-&  -& \checkmark & \textbf{87.50} \\ 
    \bottomrule
    \end{tabular}
    \caption{Affordance segmentation comparative with the state-of-the-art in the UMD dataset.}
    \label{tab:UMD_sota}
\end{table}


\begin{table*}[ht]
\centering
\resizebox{\textwidth}{!}{
\begin{tabular}{c|ccc|ccccccc}

\multicolumn{1}{c}{} & \multicolumn{1}{c}{} & \multicolumn{1}{c}{}  & \multicolumn{1}{c}{} &  \multicolumn{7}{c}{Ours}\\
 \cmidrule(lr){5-11}& & & &  Rx101&Rx101& Swin-T& Swin-T & Swin-T& Swin-T &Swin-T\\ 
 &  \cite{nguyen2016detecting}&  \cite{nguyen2017object}& 
    \cite{caselles2021standard}& 
     Determ.&MC-Drop & Determ.
   & 
   SnapShot Ens& 
    Deep Ens& 
   Mask Ens& 
    MC-Drop\\ 
\midrule
contain  & 66.4 & 75.6 & 83.6 &  85.8&85.6 & 87.2 & 89.4 & 89.2 & \textbf{89.5} & 88.8 \\
cut      & 60.7 & 69.9 & 84.7 &  86.4&84.6 & 86.7 & 88.8 & \textbf{89.0} & 87.7 & 88.9 \\
display  & 55.4 & 72.0 & 86.3 &  89.9&90.6 & 92.8 & \textbf{93.9} & 93.8 & 93.3 & 93.4 \\
engine   & 56.3 & 72.8 & 88.9 &  90.5&90.9 & 91.9 & 92.5 & \textbf{92.6} & 92.1 & 92.4 \\
grasp    & 59.0 & 63.7 & 71.1 &  77.5&78.2 & 85.6 & 84.5 & 85.2 & 84.6 & \textbf{88.9} \\
hit      & 60.8 & 66.6 & 92.3 &  94.3&95.4 & 94.5 & \textbf{95.5} & 95.4 & 95.4 & 95.4 \\
pound    & 54.3 & 64.1 & 81.8 &  80.5&80.7 & 85.1 & 87.2 & \textbf{87.9} & 86.9 & 87.2 \\
support  & 55.4 & 65.0 & 86.7 &  85.4&88.7 & 88.6 & 92.3 & \textbf{92.4} & \textbf{92.4} & 91.2 \\
w-grasp  & 50.7 & 67.3 & 83.7 &  84.7&86.6 & 87.0 & 90.2 & 89.9 & \textbf{90.8} & 89.5 \\ 
\midrule
Average  & 57.6 & 68.6 & 84.4 &  86.1&86.9 & 88.3 & 90.5 & \textbf{90.6} & 90.3 & 90.2 \\ 
\midrule
Nº train parameters (M) & - & - & 43.7 &  107.1&107.1 & 47.4 & 1258.2 & 1138.5 & 52.4 & 47.4 \\ 
Inf. Time (ms) & - & - & 45 &  47&1207 & 42 & 1015 & 1015 & 1015 & 1015 \\ 
\bottomrule
\end{tabular}
}
\caption{Per-class $F_\beta^w (\uparrow)$ affordance segmentation scores on the IIT-Aff test split dataset.}
\label{tab:my-table}
\end{table*}

\begin{table*}[ht]
\centering
\resizebox{\textwidth}{!}{
\begin{tabular}{c c cccc ccc}
 \multicolumn{1}{c}{} & \multicolumn{1}{c}{}  & \multicolumn{4}{c}{Calibration metrics} & \multicolumn{3}{c}{Performance metrics} \\
 \cmidrule(lr){3-6} \cmidrule(lr){7-9} 
 & & Semantic& Semantic& Spatial & Spatial & $F_{\beta}^w$ & PMQ & pPMQ \\ 
Backbone & Method& AUSE $(\downarrow)$ & ECE $(\downarrow)$ & AUSE $(\downarrow)$ & ECE $(\downarrow)$& score $(\uparrow)$ & $(\uparrow)$& $(\uparrow)$ \\ 
\midrule
\multirow{5}{*}{R50} 
& Deterministic & 0.308 & 0.0214 & 0.252 & 0.00483 & 84.4 & 27.7 & 55.5  \\ 
& MC-Dropout  & 0.280 & 0.0128 & \textbf{0.132} & \textbf{0.00246} & 85.9 & 41.9 & 71.1  \\ 
& Mask-Ens & 0.225 & 0.0125 & 0.186 & 0.00263 & 84.9 & 36.3 & 68.7  \\ 
& Deep-Ens & 0.241 & 0.0134 & 0.158 & 0.00257 & \textbf{87.1} & \textbf{50.5} & \textbf{71.9}  \\ 
& Snap-Ens & \textbf{0.176} & \textbf{0.0112} & 0.177 & 0.00253 & 86.3 & 40.7 & 70.5  \\ 
\midrule
\multirow{5}{*}{Swin-T} 
& Deterministic & 0.213 & 0.0146 & 0.155 & 0.00209 & 89.0 & 63.6  & 70.1 \\ 
& MC-Dropout & 0.196 & 0.0112 & 0.129 & 0.00215 & 89.6 & 37.8 & 73.2 \\ 
& Mask-Ens & 0.166 & 0.0089 & 0.102 & 0.00174 & 89.7 & 46.1 & 75.6  \\ 
& Deep-Ens & \textbf{0.136} & 0.0105 & 0.082  & 0.00167 & \textbf{90.6} & 57.5 & 77.3 \\ 
& Snap-Ens & 0.145 & \textbf{0.0084} & \textbf{0.080} & \textbf{0.00159} & 90.5 & \textbf{60.3} & \textbf{77.5}  \\ 
\bottomrule
\end{tabular}
}
\caption{\textbf{Ablation study on deterministic and Bayesian deep learning techniques.}
We compare final performance and uncertainty calibration across the different uncertainty estimation methods (MC-Dropout, Mask-Ens, Deep-Ens, Snap-Ens) and the respective deterministic version. We also report results for two different backbone configurations Resnet-50 and Swin-T.}
\label{tab:general_Aff_performance}
\end{table*}

\section{Results}

\subsection{Comparative with the state-of-the-art}

We first compare our model with the state-of-the-art for the IIT-Aff and UMD datasets in Tables \ref{tab:IIT_Aff_sota} and \ref{tab:UMD_sota}, respectively.
Our approach demonstrates significant improvements on IIT-Aff, where our best model version (Mask-Ens) achieves 90.65 $F_\beta^w$, outperforming the previous Mask-RCNN \cite{caselles2021standard} (84.40 $F_\beta^w$) by +6.1 percentage points (p.p). This notable gain highlights the superior generalization capabilities of our model in robotics and real-world scenarios, which are the primary On the UMD dataset, the Bayesian model achieves a competitive performance of 87.50 $F_\beta^w$ as shows Table \ref{tab:UMD_sota}. As UMD comprises isolated objects, model performance tends to saturate, resulting in smaller performance gains. Nevertheless, the Bayesian approach consistently outperforms the deterministic variant (86.03 $F_\beta^w$) in the UMD dataset, highlighting the benefits of aggregating multiple detections.

Following, Table \ref{tab:my-table} presents a detailed per-class comparative on the IIT-Aff dataset against previous works across different versions of our approach. 
Our Swin-T Deep-Ens achieves a 90.6 $\%$ $F_{\beta}^w$ score, with particularly strong results on the \textit{cut} 89.0 $\%$ $F_{\beta}^w$, \textit{engine} 92.6  $\%$ $F_{\beta}^w$, \textit{pound} 87.9 $\%$ $F_{\beta}^w$ and \textit{support} 92.4 $\%$ $F_{\beta}^w$ classes. 
Comparing the model size and the inference speed, the key advantage of MC-Drop and Mask-Ens is that they require training only a single model, resulting in a similar number of parameters as \cite{caselles2021standard}, while offering improved segmentation performance and uncertainty quantification. However, the computational cost of Bayesian methods is reflected in the inference time, which increases linearly with the number of samples.

\subsection{Ablation study}

\paragraph{Bayesian vs. Deterministic methods}
Table \ref{tab:general_Aff_performance} compares the differences between Bayesian and deterministic methods. \textcolor{black}{Bayesian models achieve comparable performance in affordance segmentation while consistently exhibiting lower calibration errors. For example, when comparing Snapshot Ensembles with the deterministic Swin-T model, the $F_{\beta}^w$ score is similar (89.0 vs. 90.5), whereas the Bayesian configuration substantially reduces calibration error, achieving lower Semantic AUSE (0.145 vs.\ 0.213) and Spatial AUSE (0.080 vs. 0.155). 
The improvement in the calibration metrics is also consistent across different backbone configurations.}
The global consensus of multiple networks makes Bayesian models outperform the respective deterministic model due to a better generalization and mask refinement. Probabilistic models produce better calibrated probabilities, both at the semantic vectors or the binary masks, which means that the model is less overconfident and that the confidence of the probability is closer to the accuracy of the predictions.  

\paragraph{Comparative of Bayesian methods}
Table \ref{tab:general_Aff_performance} also compares different Bayesian methods for uncertainty estimation.
Ensemble methods \cite{lakshminarayanan2017simple, huang2017snapshot} achieve superior segmentation results and exhibit lower calibration errors compared to MC-Drop or Mask-Ens.
The Swin-T Deep-Ens and Swin-T Snap-Ens configurations achieve the highest performance with lower calibration errors, which we attribute to their capacity to capture diverse local optima and effectively represent the multi-modal nature of the underlying distribution.
However, this improvement comes at the cost of a linear increase in training parameters, as Table \ref{tab:general_Aff_performance} shows.
\textcolor{black}{Deep Ensembles require training $M$ independent models, which significantly increases computational cost and memory usage.}
Notably, while Deep-Ens requires training $M$ independent models, Snap-Ens mitigates computational overhead by extending the training of a single model through cyclical learning rate adjustments.
\textcolor{black}{Snapshot Ensembles therefore provide a favorable trade-off between calibration quality and computational efficiency.}
In comparison, Mask-Ens outperforms MC-Drop in calibration, benefiting from reduced correlation among the sampling masks.
\textcolor{black}{Nevertheless, its uncertainty estimates remain less expressive than those produced by full ensemble approaches.}

Table~\ref{tab:general_Aff_performance} also provides a comparison among different Bayesian uncertainty estimation techniques.
\textcolor{black}{Ensemble-based methods consistently outperform single-model sampling approaches in both calibration and segmentation quality.}
In particular, Deep Ensembles and Snapshot Ensembles achieve the lowest calibration errors and highest $F_{\beta}^w$ scores across both ResNet-50 and Swin-T backbones, reflecting their ability to capture diverse hypotheses of the underlying predictive distribution.

\begin{table}[t]
\centering
\resizebox{\textwidth}{!}{
\begin{tabular}{lcccccc}

  \multicolumn{1}{c}{}  & \multicolumn{4}{c}{Calibration metrics} & \multicolumn{2}{c}{Performance metrics} \\
 \cmidrule(lr){2-5} \cmidrule(lr){6-7} 
 & Semantic& Semantic& Spatial & Spatial & $F_{\beta}^w$ & pPMQ \\  Method& AUSE $(\downarrow)$ & ECE $(\downarrow)$ & AUSE $(\downarrow)$ & ECE $(\downarrow)$& score $(\uparrow)$ & $(\uparrow)$ \\
\hline
Det. R50        & 0.308 & 0.0214 & 0.251 & 0.00482 & 84.4 & 55.5 \\
Det. Rx101      & 0.223 & 0.0187 & \textbf{0.146} & 0.00382 & 85.6 & 62.5 \\
Det. Swin-T    & \textbf{0.213} & \textbf{0.0146} & 0.155 & \textbf{0.00208} & \textbf{89.0} & \textbf{63.6} \\
\hline
Snap-Ens R50  & 0.176 & 0.0121 & 0.176 & 0.00252 & 86.3 & 70.5 \\
Snap-Ens Rx101& 0.160 & 0.0139 & 0.105 & 0.00222 & 87.4 & 76.4 \\
Snap-Ens Swin-T & \textbf{0.145} & \textbf{0.0084} &\textbf{ 0.079} & \textbf{0.00159} & \textbf{90.5} & \textbf{77.5} \\
\hline
\end{tabular}
}
\caption{\textbf{\textcolor{black}{Ablation study on backbone architectures.}}
\textcolor{black}{We compare final performance and uncertainty calibration across three encoder backbones: ResNet-50 (R50), ResNeXt-101 (Rx101), and Swin-T. 
For each backbone, we report results for both the deterministic model and its Bayesian counterpart based on Snapshot Ensembles (Snap-Ens).}
}
\label{tab:uncertainty_metrics}
\end{table}

\paragraph{Backbone configurations}
Results in Table \ref{tab:general_Aff_performance} show that the backbone affects significantly to the performance and the uncertainty estimation: attention-based encoders obtain best affordance segmentation and show better calibration metrics.
For example, comparing the analogous Mask-Ens Resnet-50 and Swin-T configurations the differences are 84.9 vs. 89.7 $\%$ $F_{\beta}^w$, 0.225 vs. 0.166 Sem AUSE, 0.0125 vs. 0.0089 Sem ECE, 0.186 vs. 0.102 Sp AUSE and 0.00263 vs. 0.00174 Sp. ECE. 
\textcolor{black}{These results indicate that the hierarchical self-attention mechanism of Swin-T provides richer multi-scale representations, which benefit not only segmentation accuracy but also the quality of uncertainty calibration.}
\textcolor{black}{In contrast, convolutional backbones such as ResNet-50 and ResNeXt-101 exhibit higher calibration errors, suggesting more overconfident predictions, particularly along object boundaries.}
We also report in Table\ \ref{tab:general_Aff_performance} our novel PMQ metric, which evaluates the uncertainty quality of the predicted masks. The differences with the previous conference version \cite{aff_loren} are due to the computation of the spatial probability as the mean of the per-pixel masks probabilities and not as the number of times that a pixel is detected ($\mathbf{p}^{mask} > 0.5$). The improvement of Swin-T against Resnet-50 is also reflected on this metric (46.1 vs. 36.3 PMQ, 75.6 vs. 68.7, 85.9 vs. 78.9 $Q_L$, 67.9 vs. 61.9 $Q_S$ for the Mask-Ens Resnet-50 and Swin-T, respectively).
\textcolor{black}{Overall, these findings highlight that backbone choice is a key factor in uncertainty-aware affordance segmentation, with transformer-based architectures offering superior trade-offs between predictive performance and calibration quality.}

\begin{table*}[t]
\centering
\resizebox{\textwidth}{!}{
\begin{tabular}{llcccccc}
\multicolumn{2}{c}{} &
\multicolumn{4}{c}{Calibration metrics} &
\multicolumn{2}{c}{Performance metrics} \\
\cmidrule(lr){3-6} \cmidrule(lr){7-8}
Method & Sampling &
Semantic  &
Semantic &
Spatial  &
Spatial &
$F_{\beta}^w$ &
pPMQ \\
 &  &
AUSE $(\downarrow)$ &
ECE $(\downarrow)$ &
AUSE $(\downarrow)$ &
ECE $(\downarrow)$ &
score $(\uparrow)$ &
$(\uparrow)$ \\
\midrule

\multirow{5}{*}{MC-Dropout} 
& ENC            & 0.156 & 0.0110& \textbf{0.098}& 0.00186& \textbf{90.0} & \textbf{55.6 }\\
& ENC + FC       & 0.196 & 0.0084& 0.121& 0.00214& 89.6 & 37.8 \\
& ENC + MASK     & \textbf{0.142} & 0.0105& 0.100& \textbf{0.00181}& 89.2 & 47.7 \\
& ENC + FC + MASK& 0.195& \textbf{0.0081}& 0.128& 0.00219& 88.3 & 34.5 \\
& FC + MASK      & 0.193& 0.0105& 0.112& 0.00221& 88.7 & 48.7 \\
\midrule

\multirow{5}{*}{Mask-Ens} 
& ENC            & 0.154& 0.0097& 0.099& 0.00165& \textbf{90.2} & 51.5 \\
& ENC + FC       & 0.165& \textbf{0.0089}& 0.101& 0.00173& 89.7 & 46.1 \\
& ENC + MASK     & \textbf{0.152}& 0.0094& \textbf{0.090}& \textbf{0.00164}& 89.3 & 46.8 \\
& ENC + FC + MASK& 0.168& \textbf{0.0089}& 0.108& 0.00191& 88.4 & 42.5 \\
& FC + MASK      & 0.197& 0.0157& 0.128& 0.00291& 88.3 & \textbf{57.3} \\
\bottomrule
\end{tabular}
}
\caption{\textbf{\textcolor{black}{Ablation study of sampling locations for uncertainty estimation.}}
\textcolor{black}{We report calibration (AUSE, ECE) and performance ($F_{\beta}^w$, pPMQ) metrics for different sampling configurations applied to Swin-T using MC Dropout and Mask Ensembles.}}
\label{tab:sampling_ablation}
\end{table*}

\paragraph{Disposition of sampling layers}
We conduct an ablation study on the placement of MC-Dropout sampling layers in Table \ref{tab:sampling_ablation}. 
When sampling layers are positioned in the encoder (ENC), the model learns a probabilistic latent space that propagates through the architecture, leading to better-calibrated probability estimates. 
\textcolor{black}{This indicates that injecting stochasticity early in the feature extraction process allows uncertainty to be encoded at multiple semantic levels, rather than being constrained to the final predictions.}
In contrast, when sampling layers are applied only in the final layers (FC + M), enforcing probabilistic outputs at this stage results in less consistent and less well-calibrated predictions.
\textcolor{black}{This behavior suggests that late-stage sampling is insufficient to capture epistemic uncertainty arising from ambiguous visual representations.}
Additionally, we observe that ENC + M provides better calibration for binary masks, whereas ENC + FC improves the calibration of semantic probabilities due to the proximity of the dropout layers. Comparing Swin-T MC-Dropout with Mask-Ens, the results indicate that Mask-Ens slightly outperforms MC-Dropout in both performance and calibration. This improvement is attributed to the lower correlation among its dropout layers, allowing Mask-Ens to generalize more effectively.

\begin{figure}[h]
  \centering
  \includegraphics[width=0.48\linewidth]{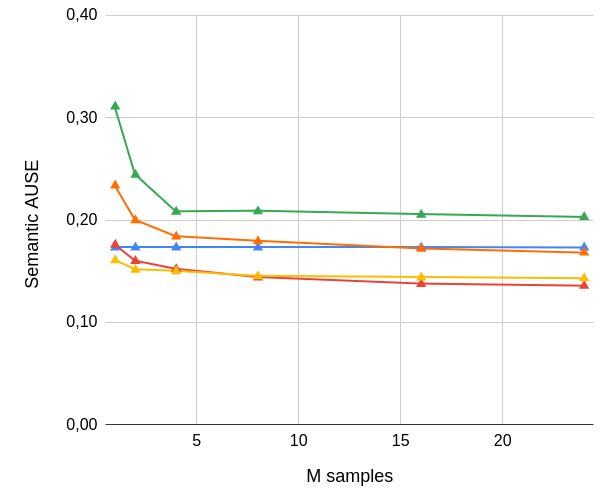}
  \includegraphics[width=0.48\linewidth]{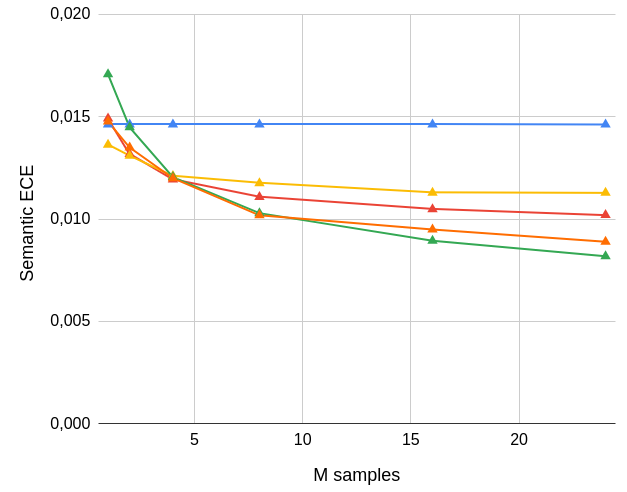} \\
  \includegraphics[width=0.48\linewidth]{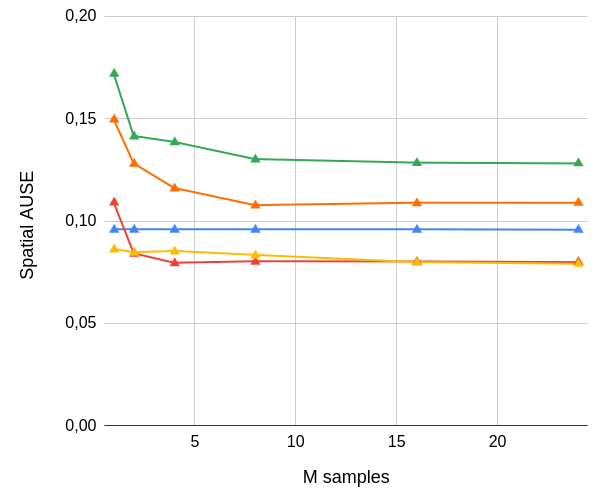}
  \includegraphics[width=0.48\linewidth]{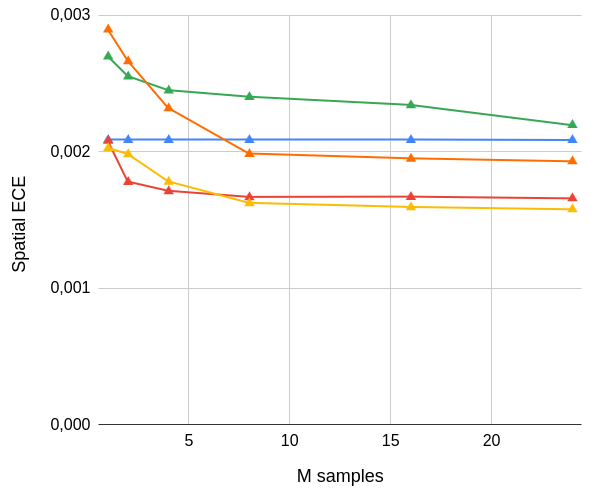} 

  \includegraphics[width=0.5\linewidth]{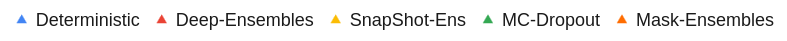}\\

    \caption{Evolution of the calibration metrics with the number of samples $M$ for different uncertainty extraction versions on the Swin-T encoder.
    }
  \label{fig:calibration_Ev}
\end{figure}

\begin{figure}[h]
  \centering
  \includegraphics[width=0.99\linewidth]{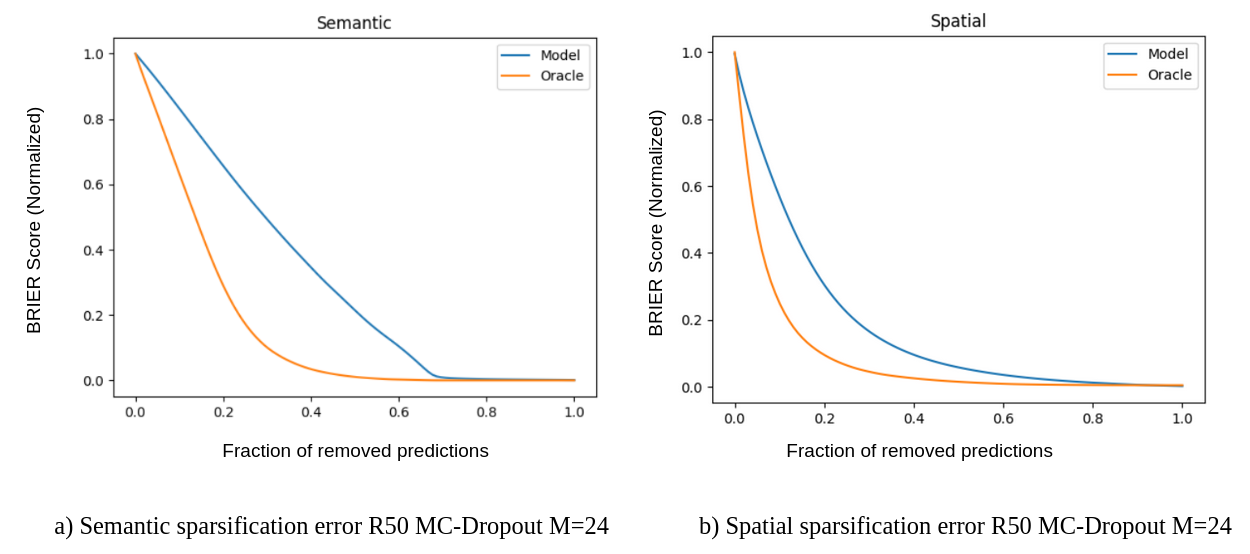}\\
  \includegraphics[width=0.99\linewidth]{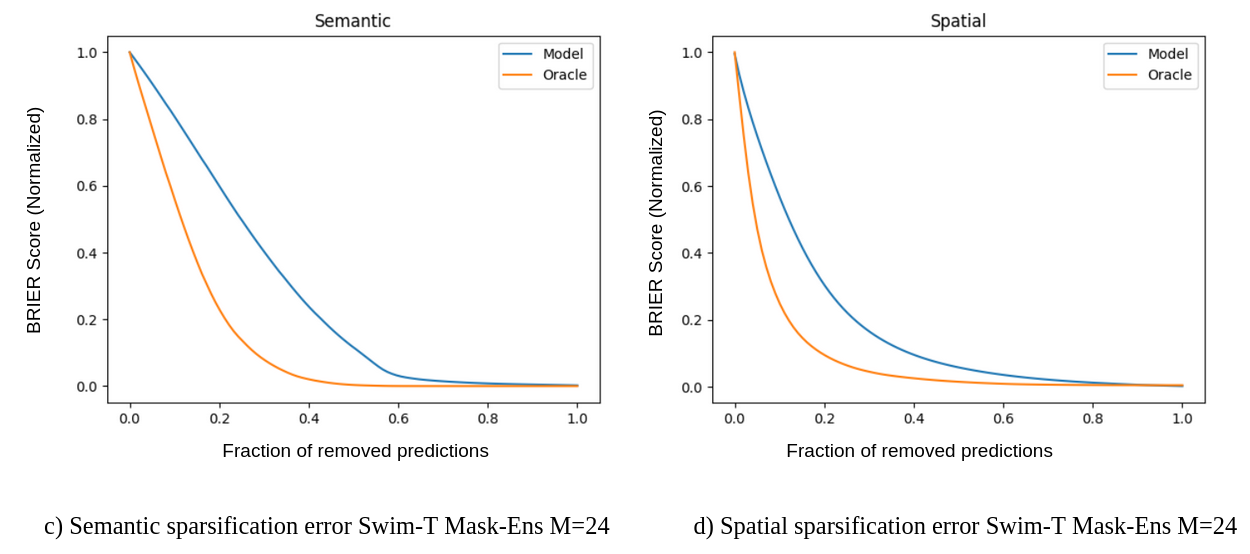}

    \caption{Sparsification error curves for the semantic and spatial probabilities for Swin-T}
  \label{fig:calibration_spars}
\end{figure}

\paragraph{Number of samples}
We also present the evolution of calibration metrics as a function of the number of forward passes in Figure \ref{fig:calibration_Ev}. The curves indicate rapid convergence within the first $M=8$ samples, followed by a plateau phase. 
Additionally, the sparsification error curves in Figure \ref{fig:calibration_spars} demonstrate that spatial uncertainty more accurately approximates the true error than semantic uncertainty, as evidenced by the smaller area difference between the corresponding curves in the spatial case. The reduction in calibration errors further confirms that the estimated probabilities more closely align with the true probabilities.

\begin{figure*}
\centering

\includegraphics[width=0.03\textwidth]{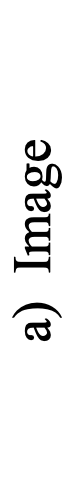} 
\includegraphics[width=0.19\textwidth]{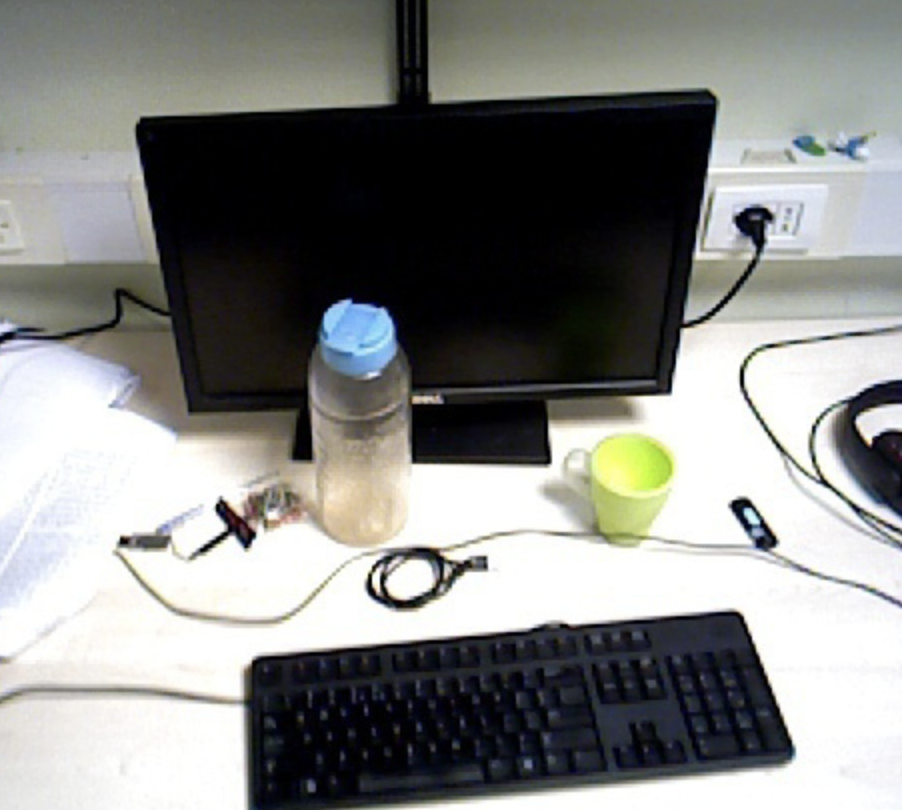} 
\includegraphics[width=0.18\textwidth]{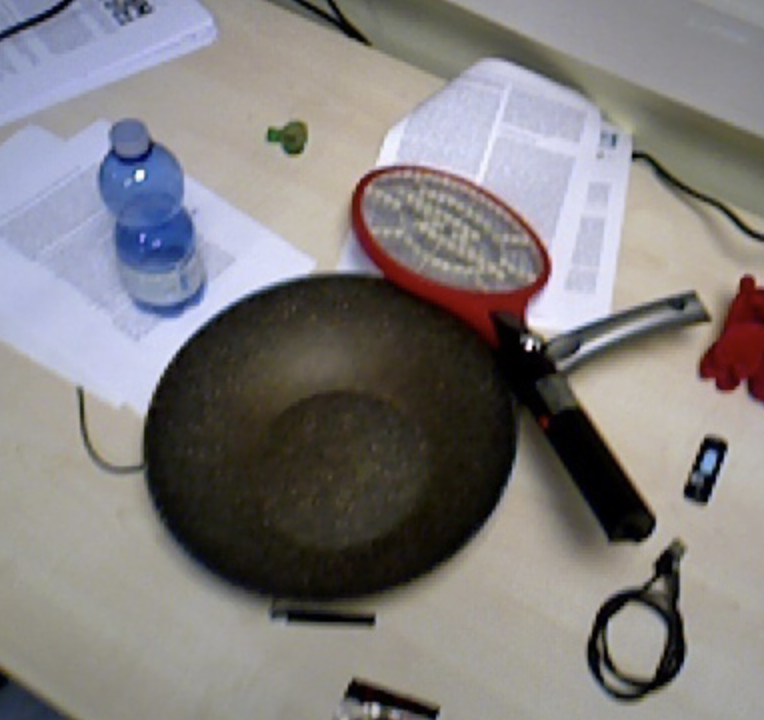} 
\includegraphics[width=0.21\textwidth]{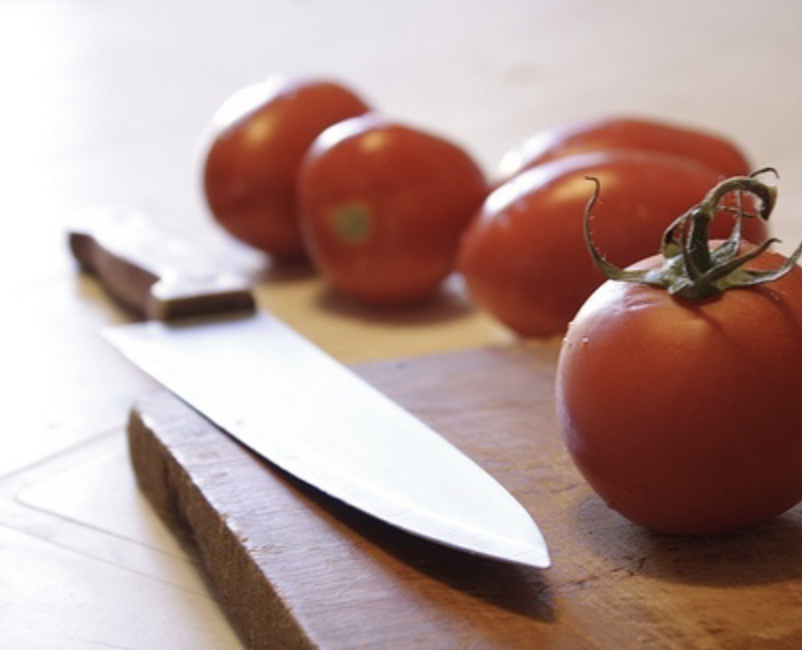} 
\includegraphics[width=0.2\textwidth]{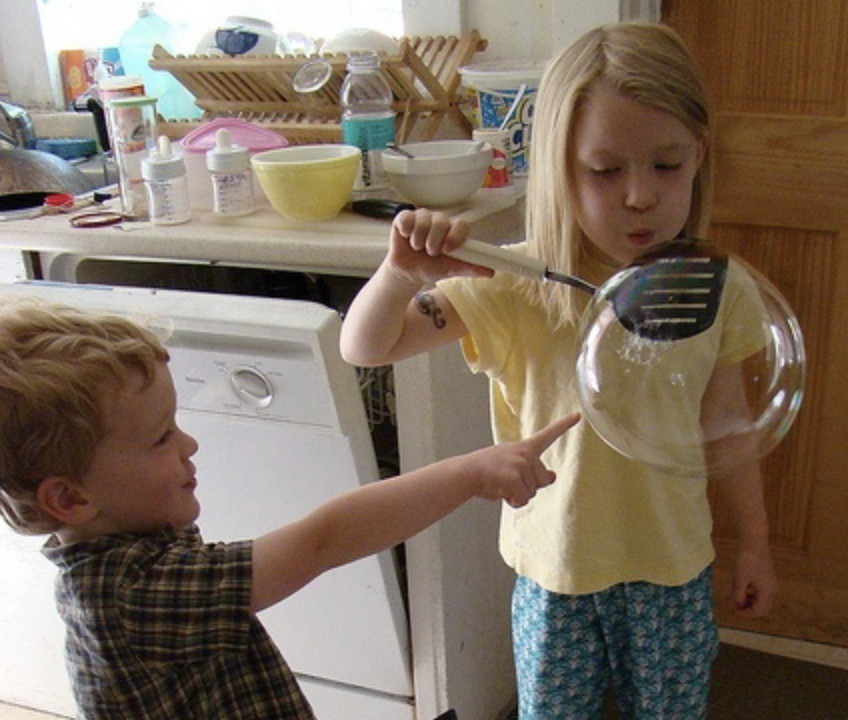} 
\includegraphics[width=0.13\textwidth]{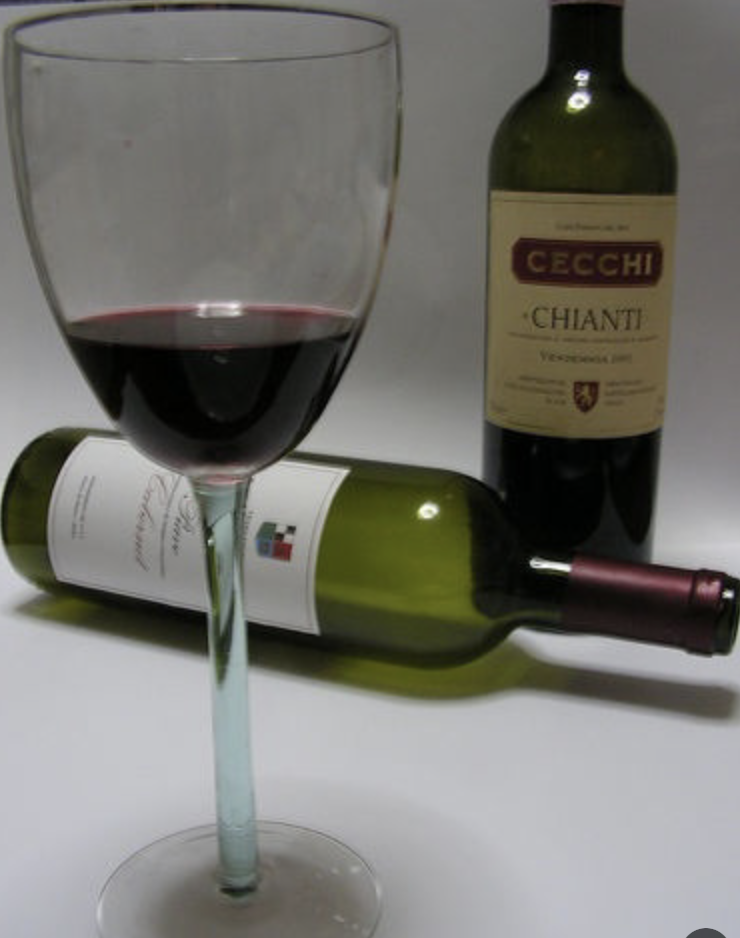}\\

\includegraphics[width=0.03\textwidth]{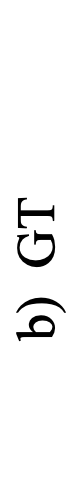} 
\includegraphics[width=0.19\textwidth]{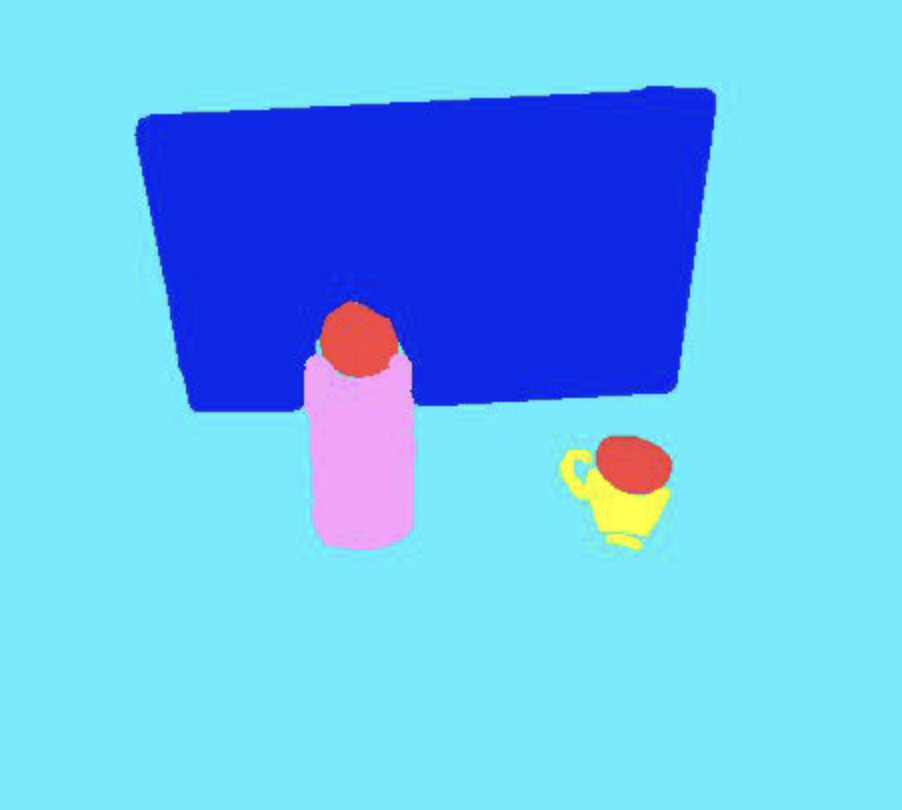} 
\includegraphics[width=0.18\textwidth]{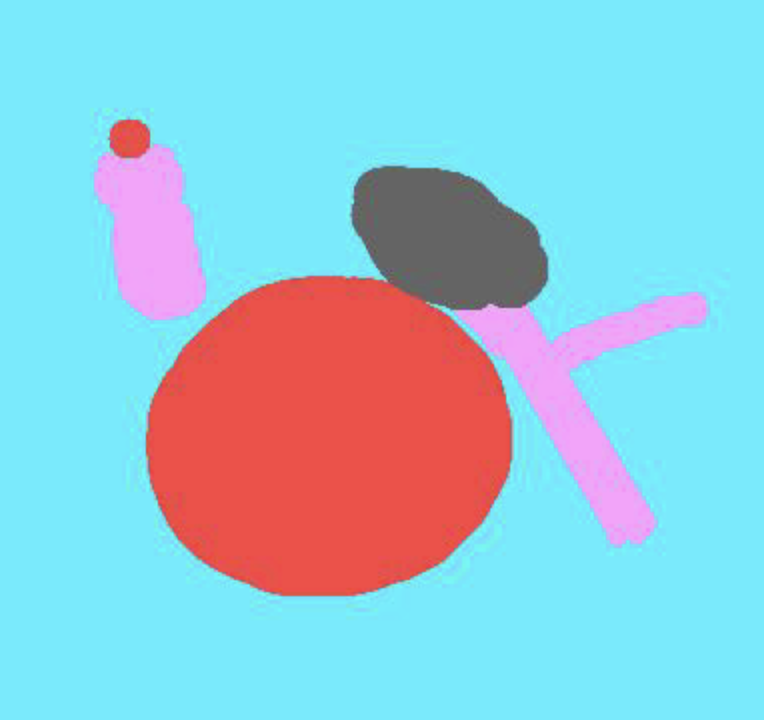} 
\includegraphics[width=0.21\textwidth]{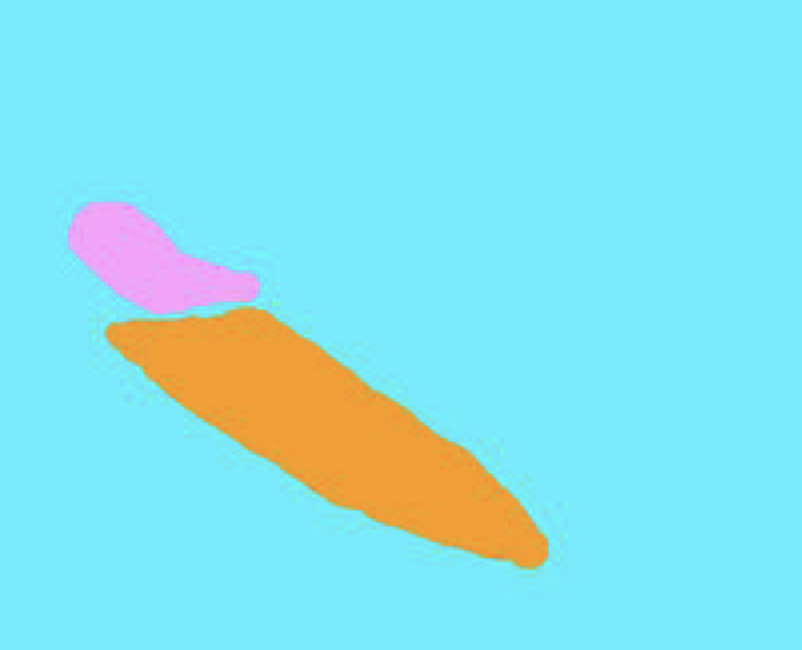} 
\includegraphics[width=0.2\textwidth]{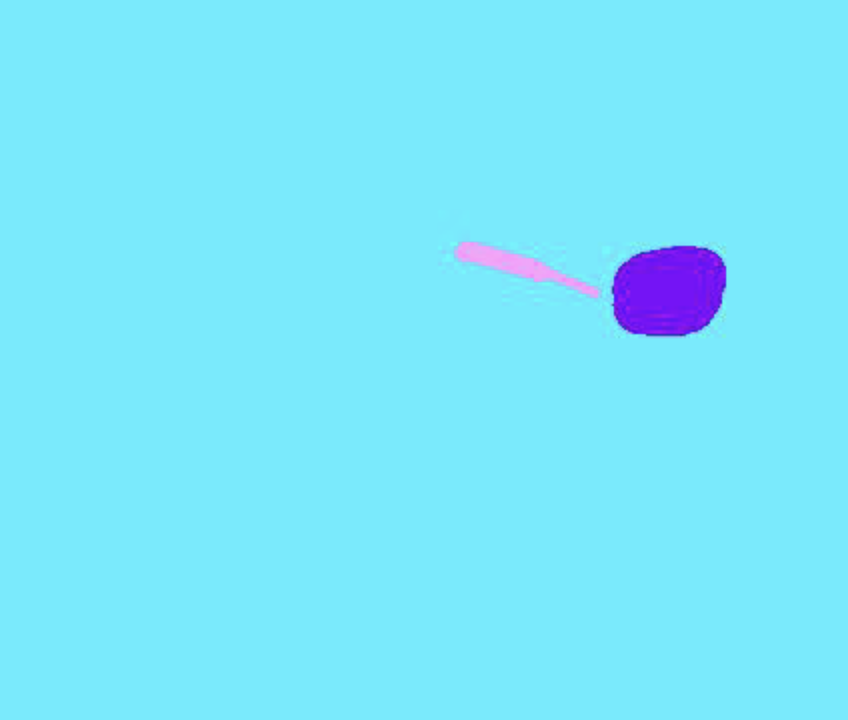} 
\includegraphics[width=0.13\textwidth]{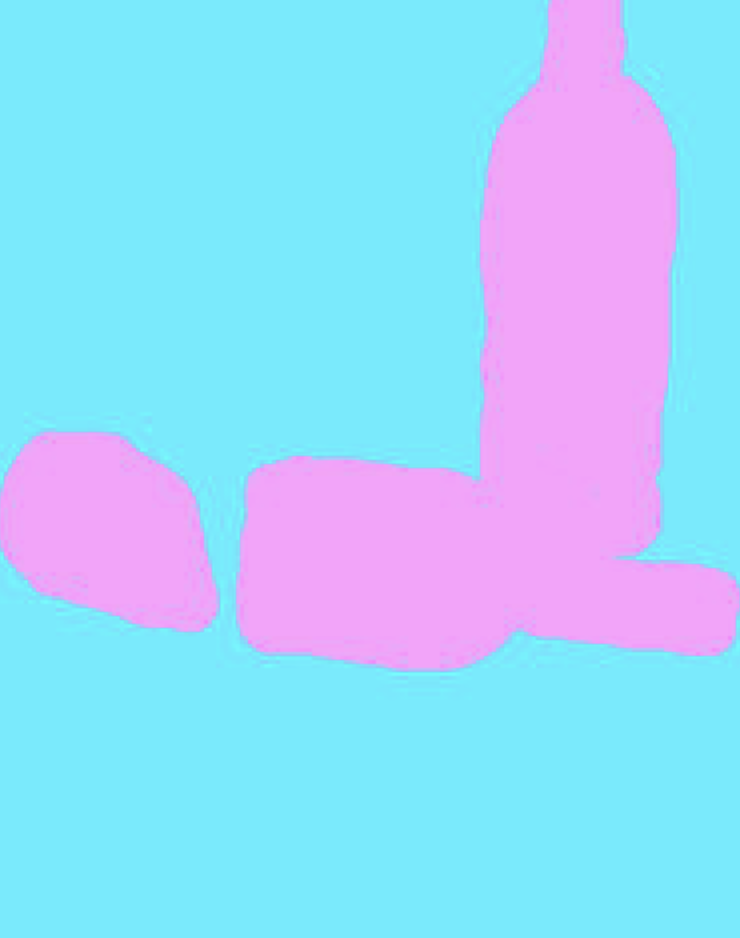}\\

\includegraphics[width=0.03\textwidth]{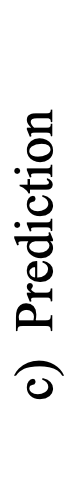} 
\includegraphics[width=0.19\textwidth]{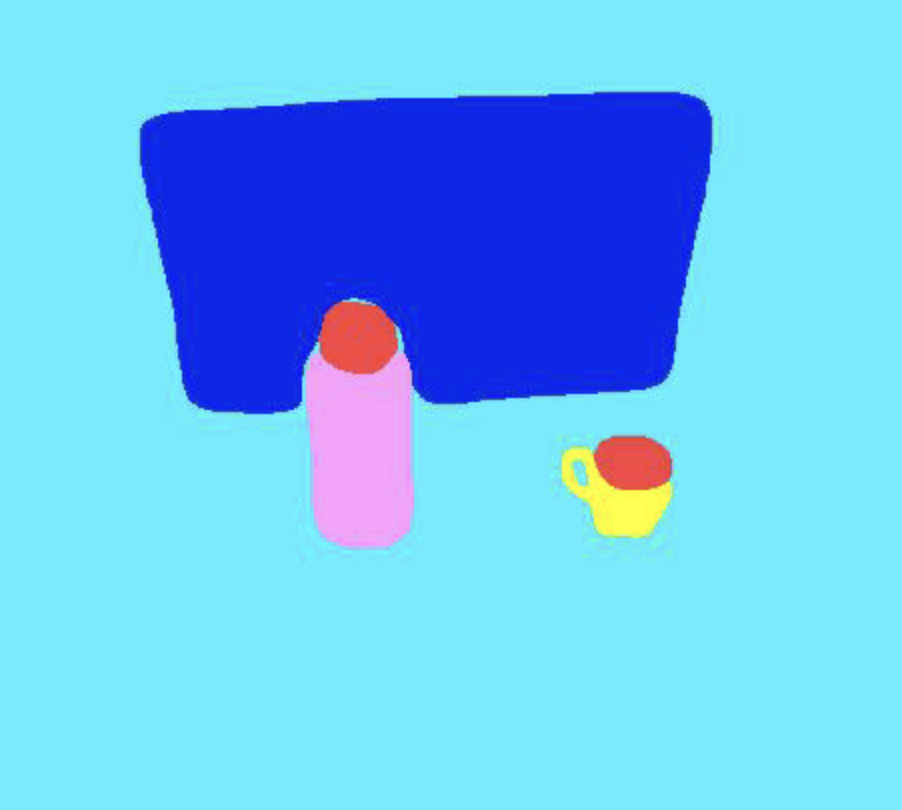} 
\includegraphics[width=0.18\textwidth]{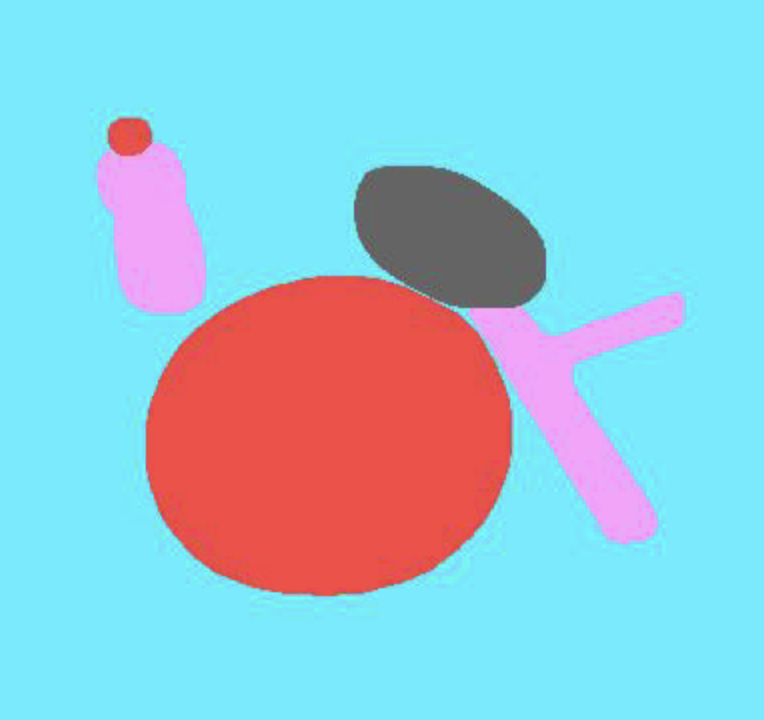} 
\includegraphics[width=0.21\textwidth]{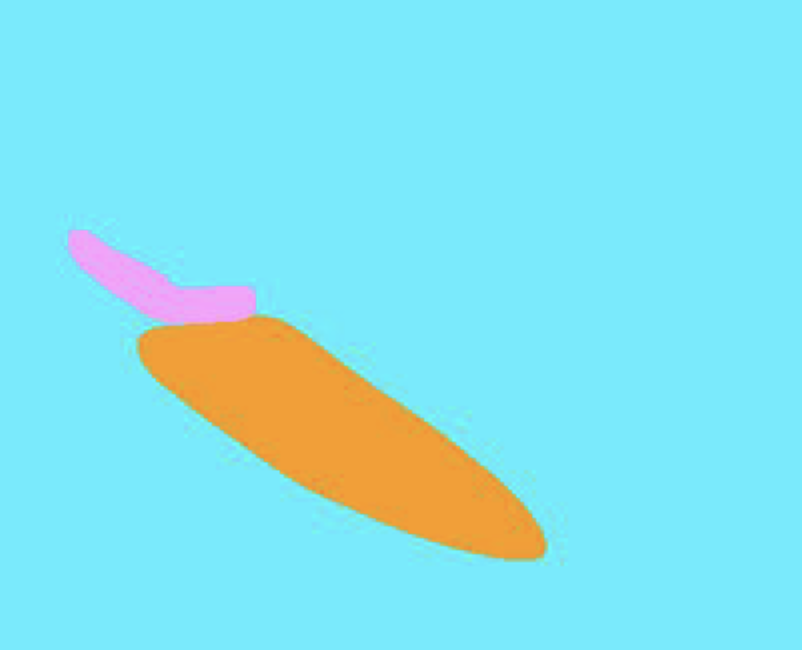} 
\includegraphics[width=0.2\textwidth]{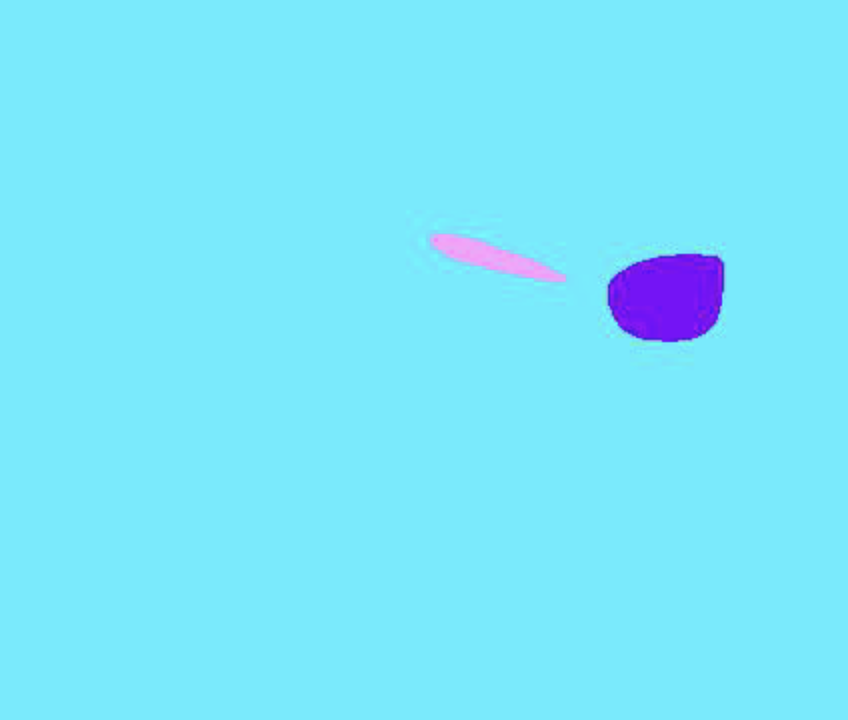} 
\includegraphics[width=0.13\textwidth]{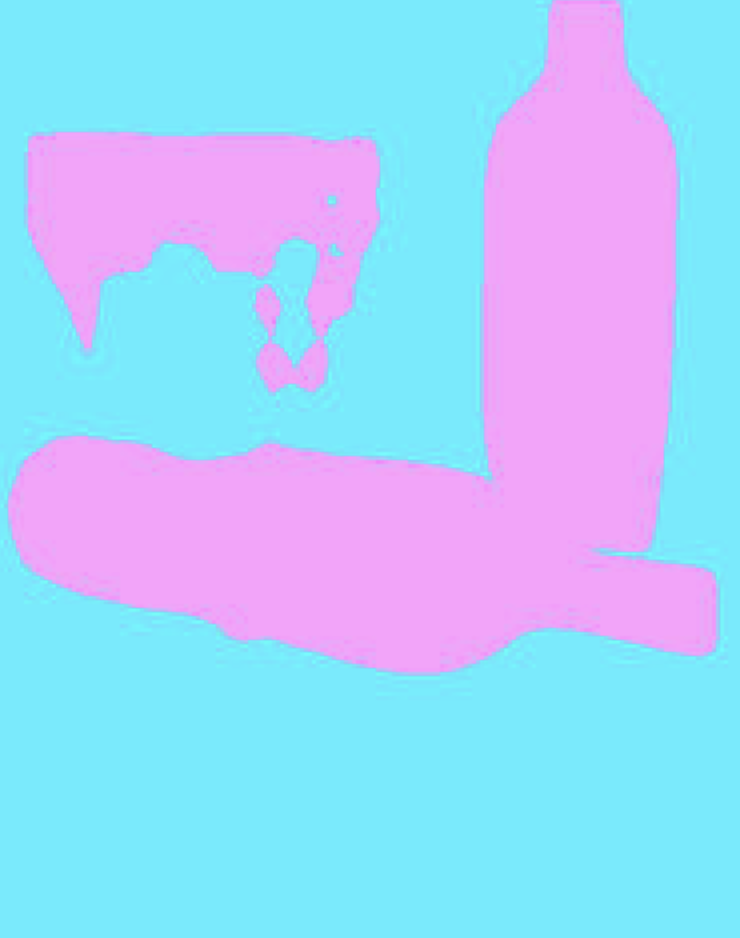}\\

\includegraphics[width=0.03\textwidth]{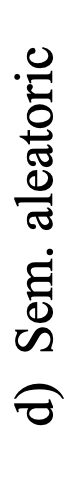}
\includegraphics[width=0.19\textwidth]{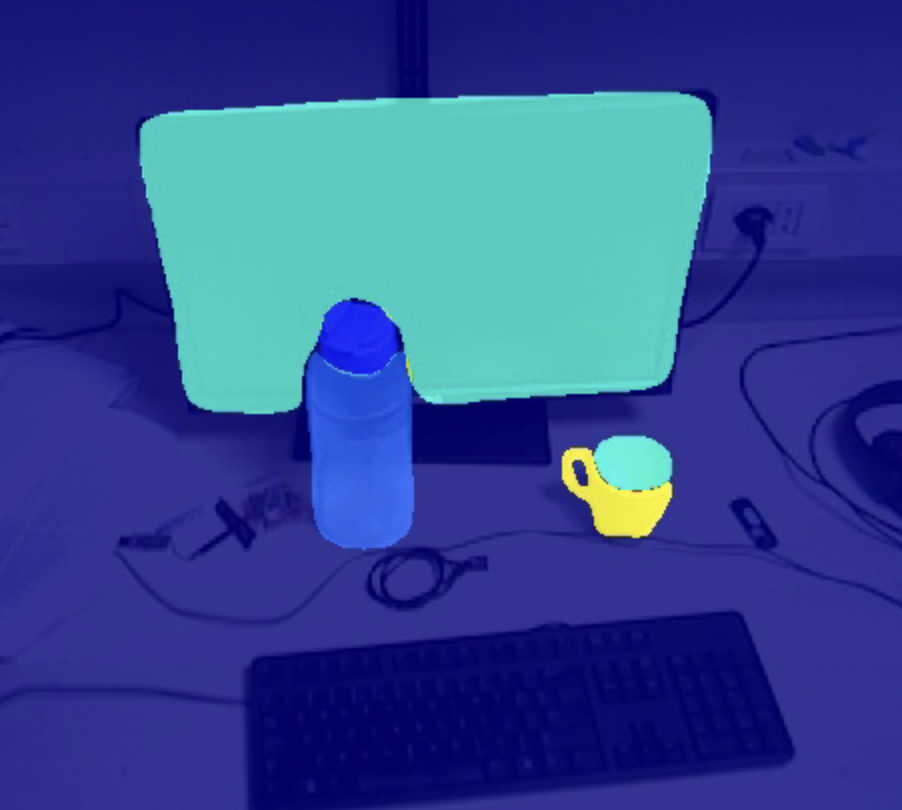} 
\includegraphics[width=0.18\textwidth]{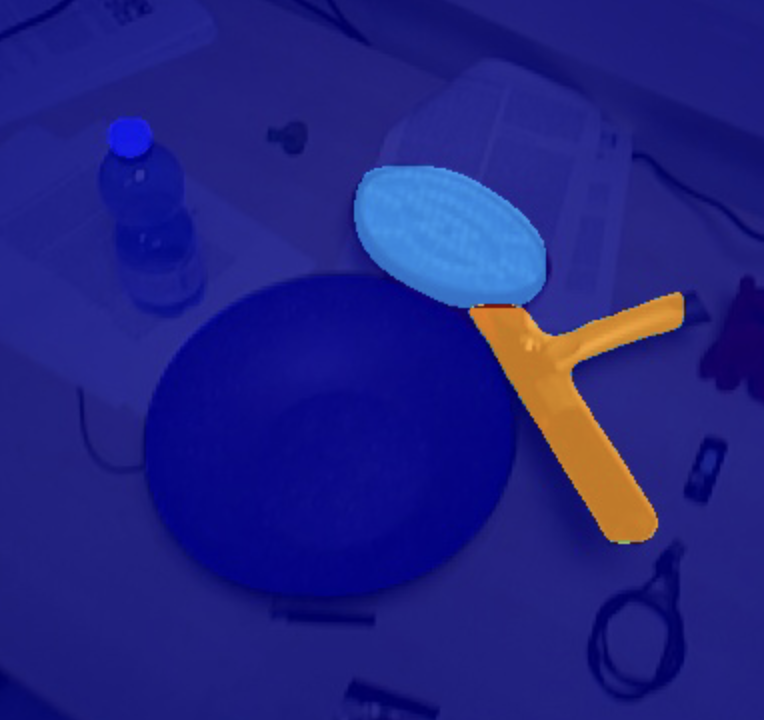} 
\includegraphics[width=0.21\textwidth]{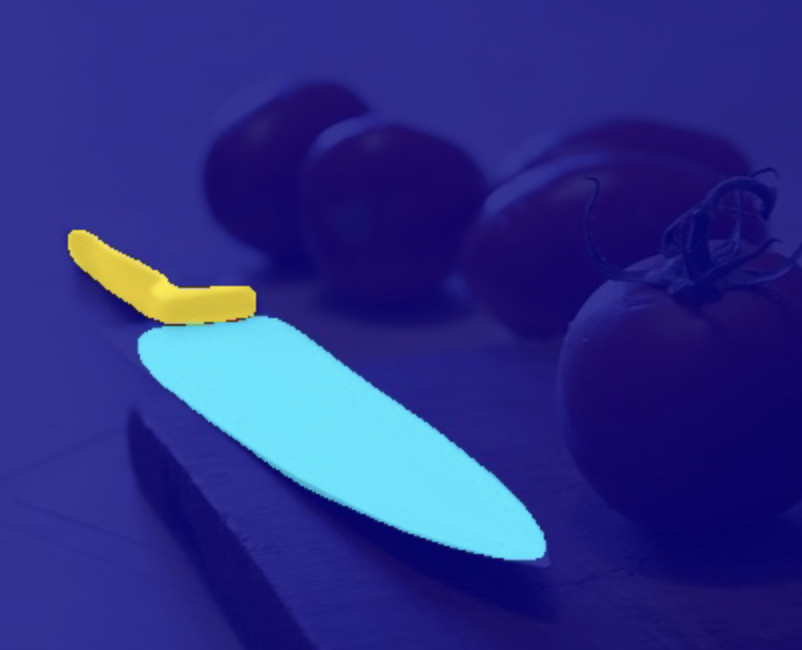} 
\includegraphics[width=0.2\textwidth]{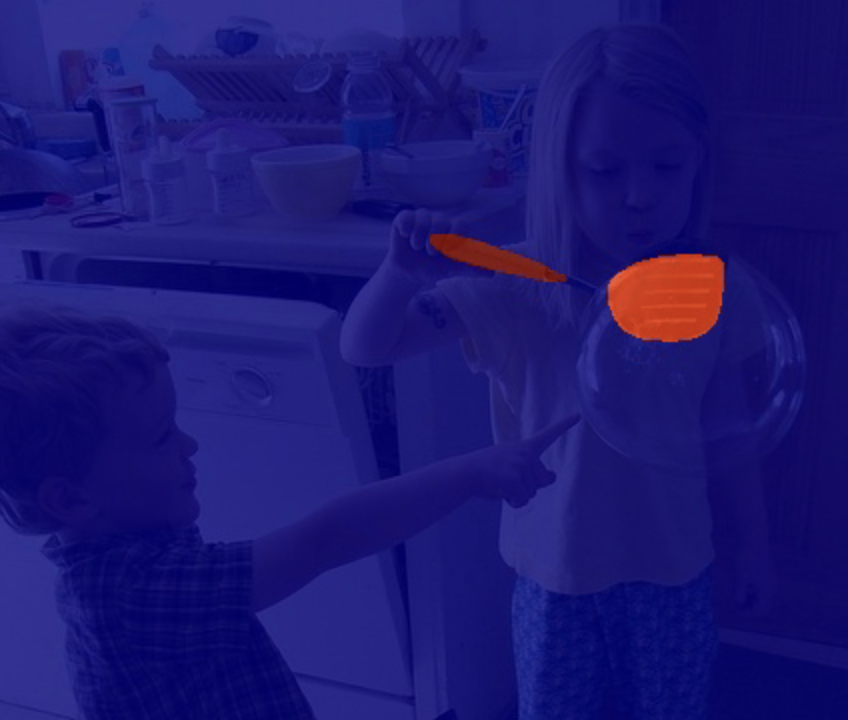} 
\includegraphics[width=0.13\textwidth]{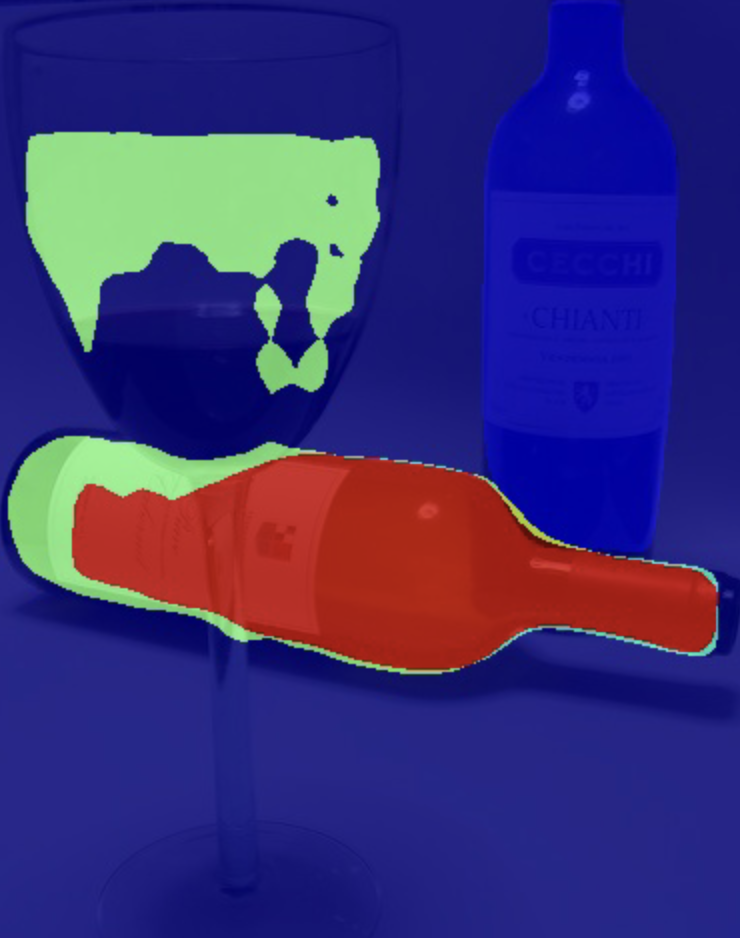}\\

\includegraphics[width=0.03\textwidth]{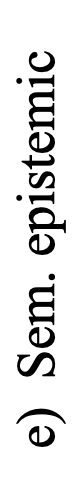}
\includegraphics[width=0.19\textwidth]{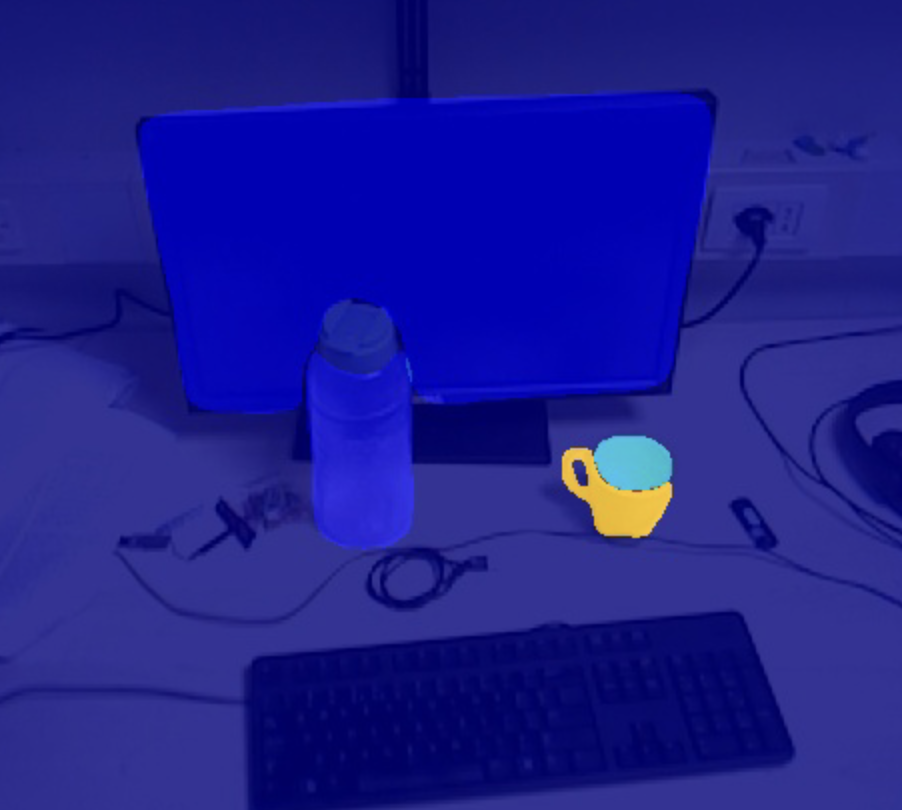} 
\includegraphics[width=0.18\textwidth]{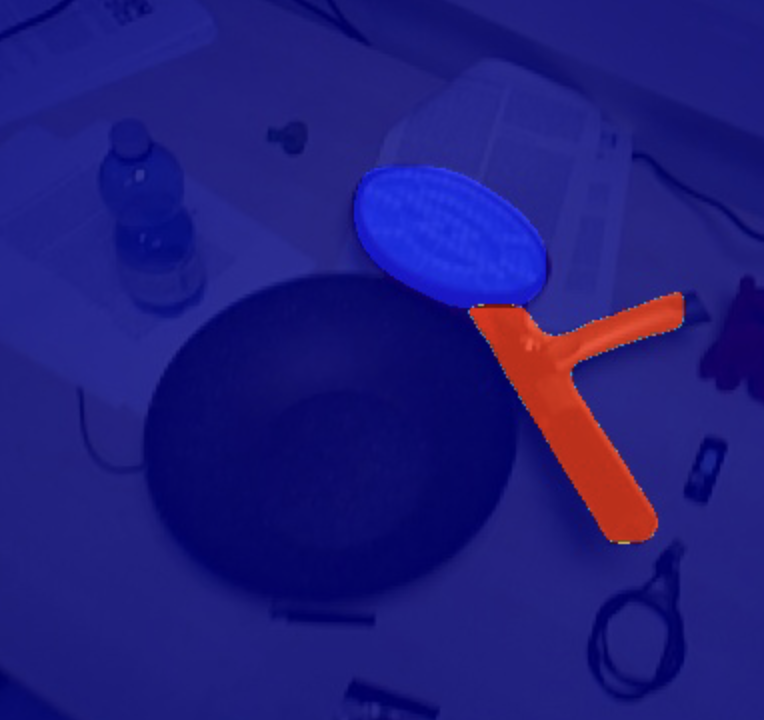} 
\includegraphics[width=0.21\textwidth]{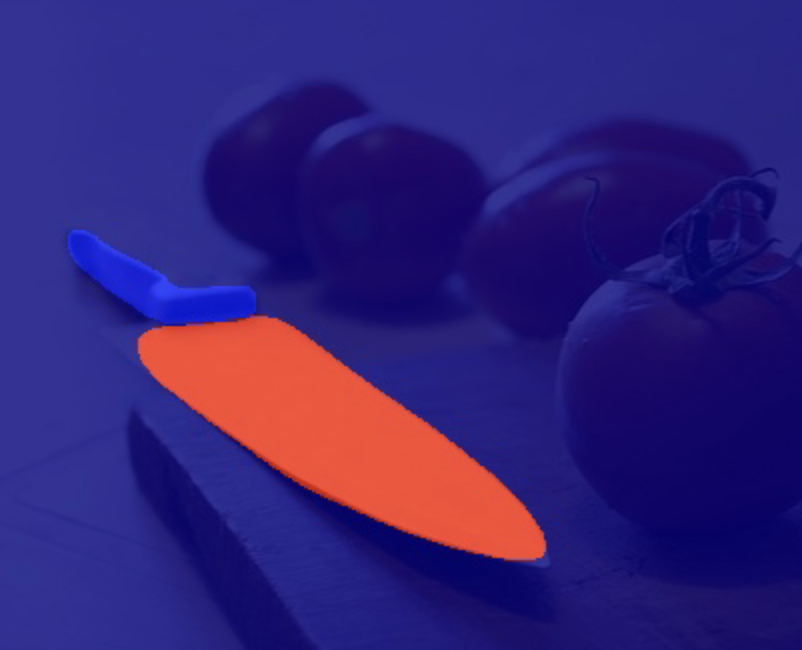} 
\includegraphics[width=0.2\textwidth]{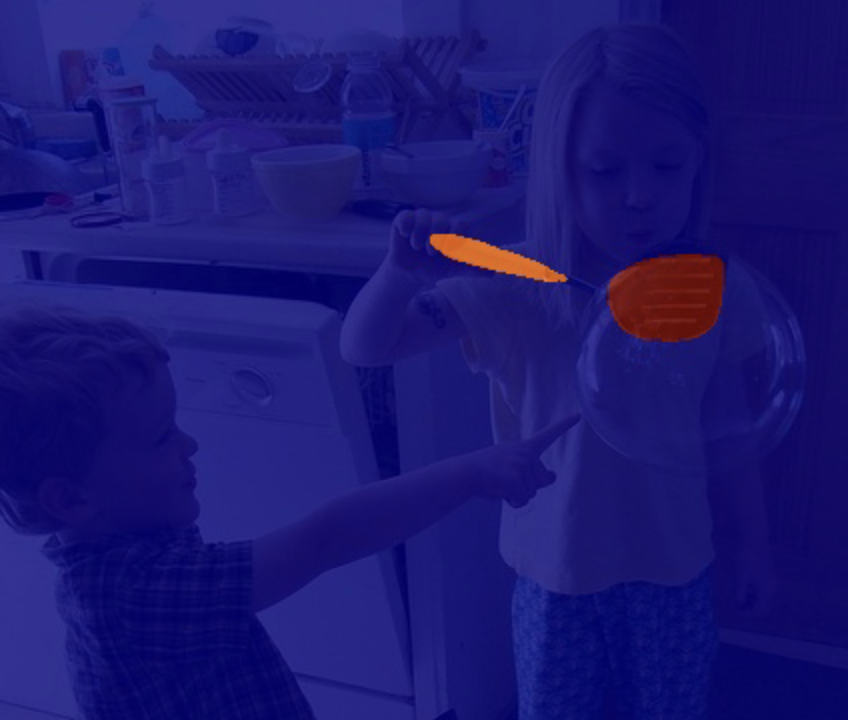} 
\includegraphics[width=0.13\textwidth]{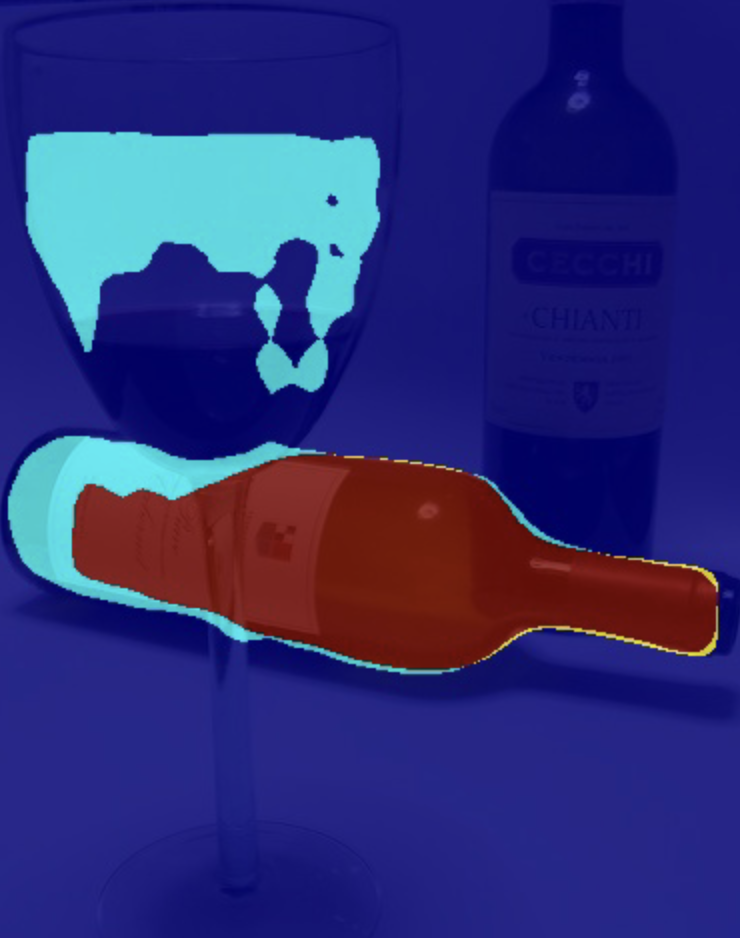}\\

\includegraphics[width=0.03\textwidth]{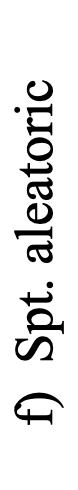}
\includegraphics[width=0.19\textwidth]{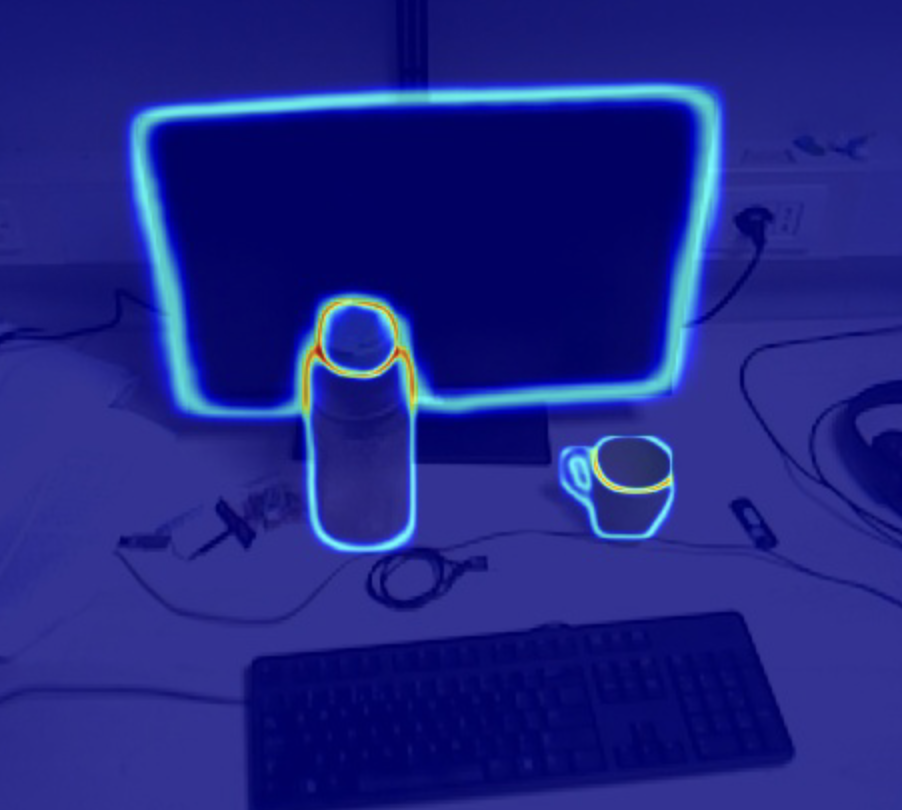} 
\includegraphics[width=0.18\textwidth]{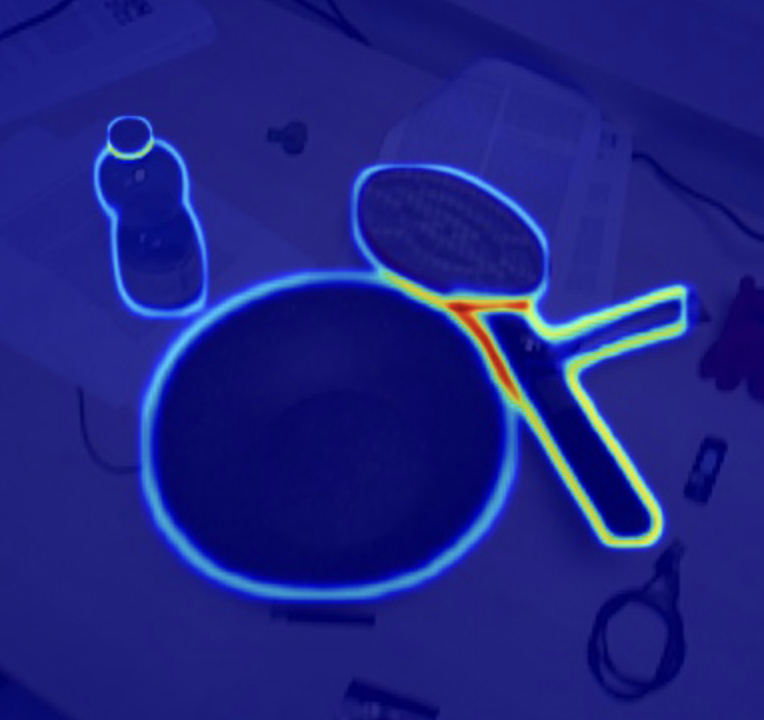} 
\includegraphics[width=0.21\textwidth]{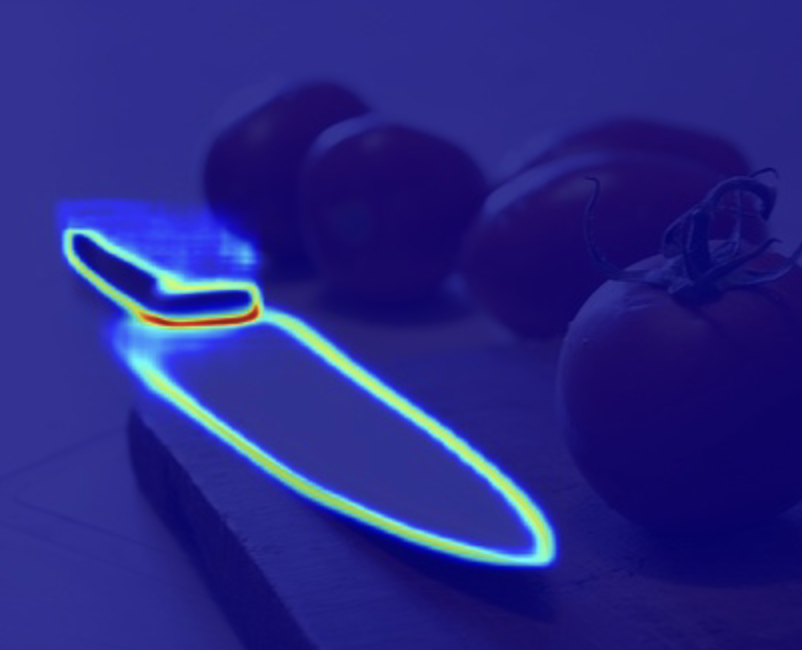} 
\includegraphics[width=0.2\textwidth]{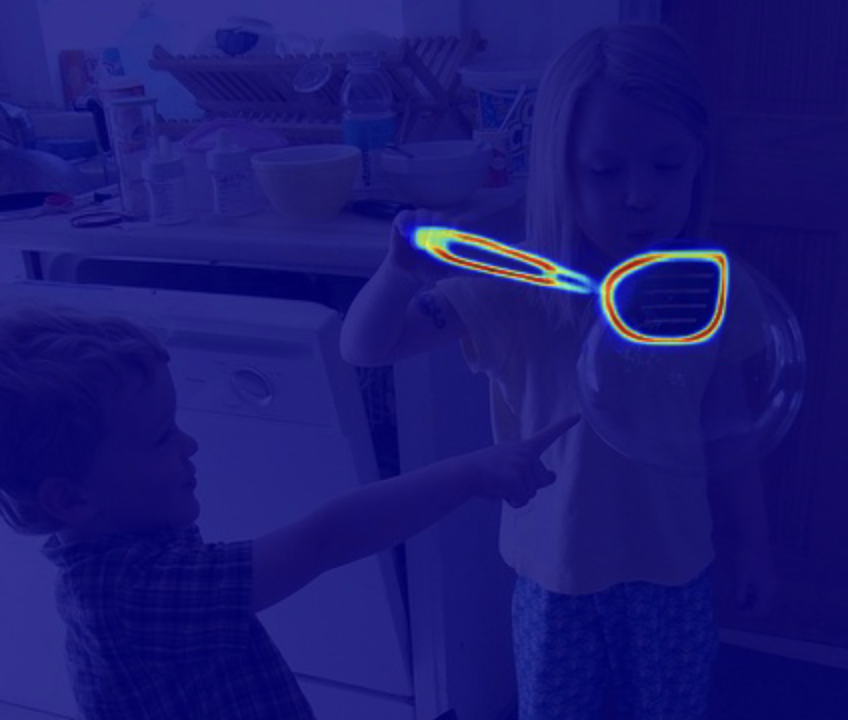} 
\includegraphics[width=0.13\textwidth]{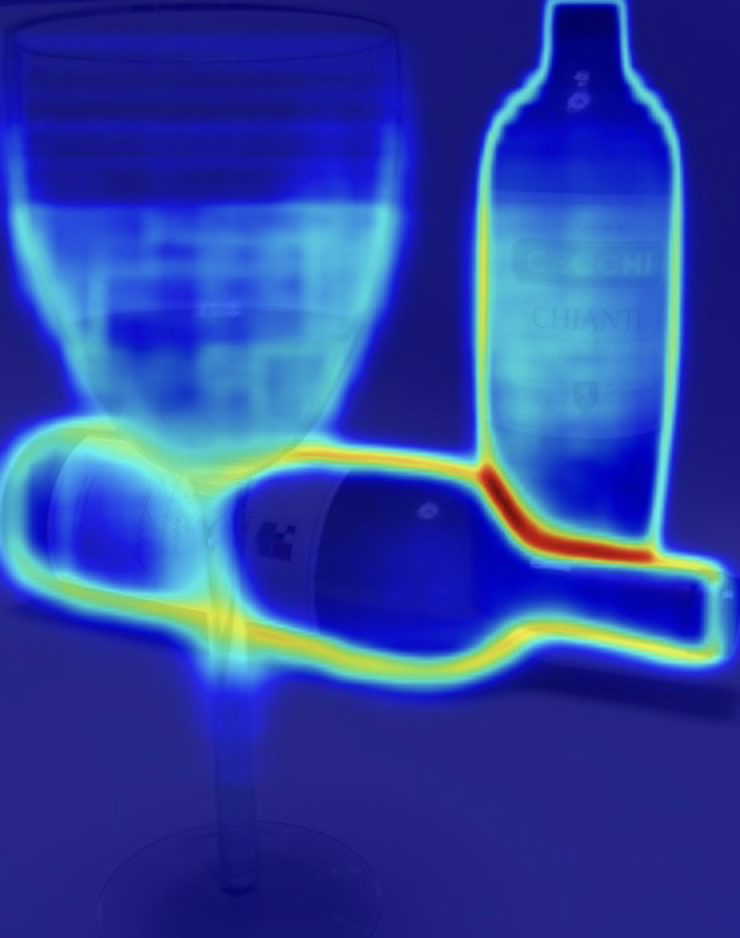}\\

\includegraphics[width=0.03\textwidth]{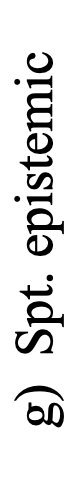}
\includegraphics[width=0.19\textwidth]{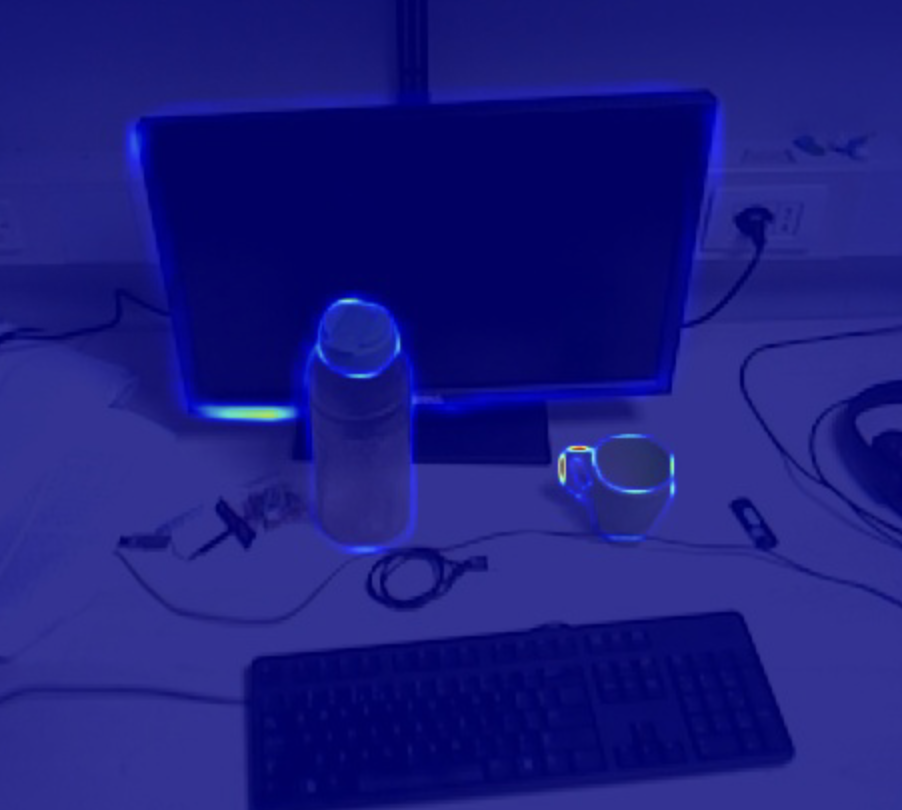} 
\includegraphics[width=0.18\textwidth]{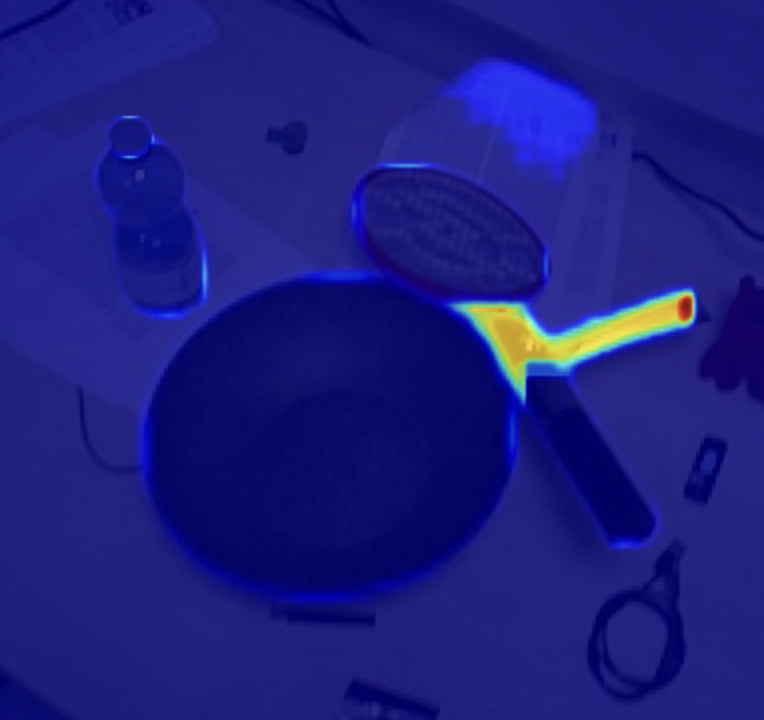} 
\includegraphics[width=0.21\textwidth]{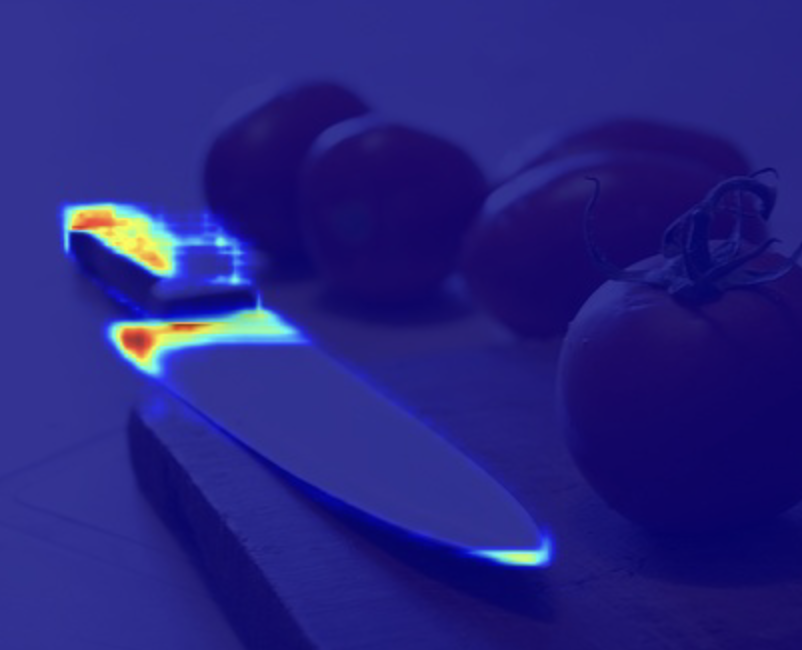} 
\includegraphics[width=0.2\textwidth]{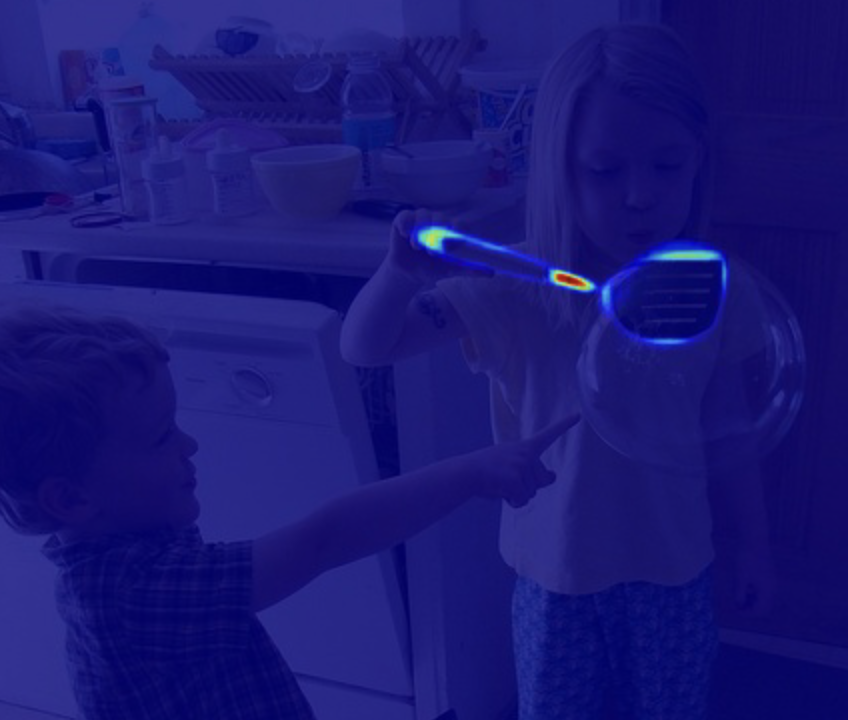} 
\includegraphics[width=0.13\textwidth]{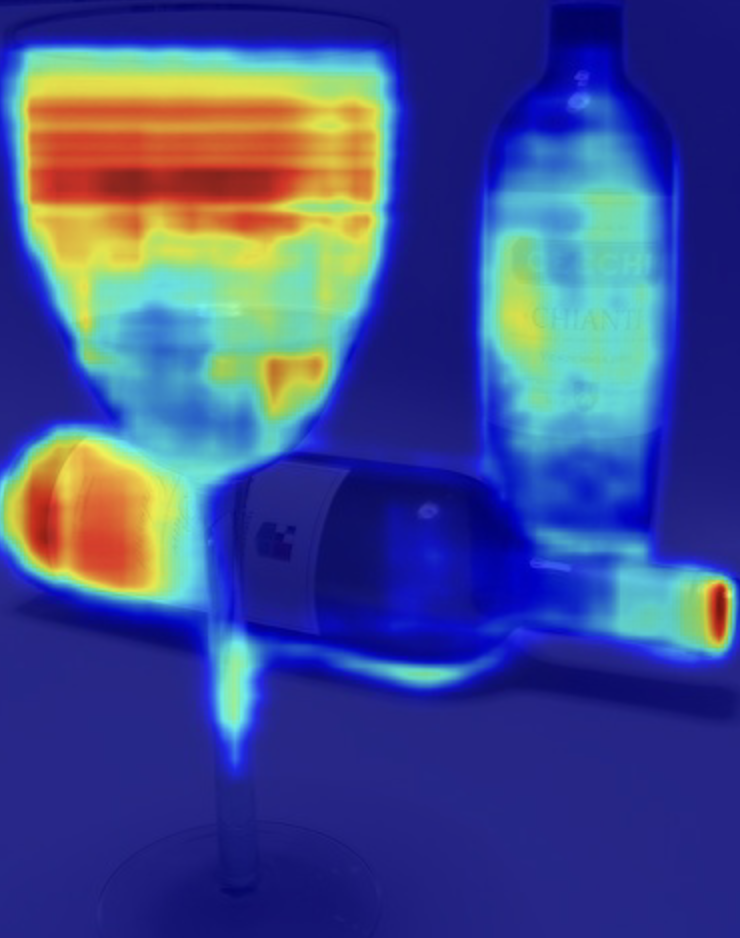}\\

\caption{Qualitative results and uncertainty prediction obtained by the Swin-T Mask-Ens Bayesian Instance segmentation model}
\end{figure*}

\subsection{Qualitative results}

We show an extensive example of the qualitative results in Figure 6. The predicted affordance maps are very close to the ground-truth labels with smooth borders as a consequence of the multiple mask averaging. While the semantic uncertainty is constant for all the predicted pixels of the object class, the spatial uncertainty is pixel-wise distributed. The spatial aleatoric uncertainty, due to the data noise, is mostly present in the object's contours. We encounter the highest values of this uncertainty at the intersection of different masks. The spatial epistemic variance reflects the model's ignorance about the detection and adds comprehensiveness when the model fails the segmentation. By \cite{Gal}, epistemic uncertainty appears in challenging pixels out-of-the-distribution. For example, in the third example, the model fails when predicting a part of the knife handle, due to the high reflections that make these pixels challenging to segment.  
\section{Conclusions}

In this work, we introduce the \emph{instance segmentation of affordances with uncertainty estimation}. We extend an attention-based backbone with different techniques for uncertainty quantification. We conducted a comparative analysis of ensemble methods (Deep-Ens \cite{lakshminarayanan2017simple} and Snap-Ens \cite{huang2017snapshot}) against sampling-based techniques (MC-Dropout \cite{gal2016dropout} and Mask-Ens \cite{durasov2021masksembles}). A detailed ablation study further examines the effects of MC-Dropout sampling layer configurations and the influence of the quantity of training parameters on model performance. We obtain a new state-of-the art at 90.6 $\% F_\beta^w$ score, which represents a significant improvement over \cite{aff_loren} and \cite{caselles2021standard}. We further extend the uncertainty quantification with the estimation of fine-grained spatial and semantic variance maps, both with the epistemic and aleatoric contributions and evaluated quantitatively with our novel PMQ metric.


\bibliographystyle{elsarticle-num-names} 
\bibliography{mybib}






\end{document}